# Trend Extrapolation for Technology Forecasting: Leveraging LSTM Neural Networks for Trend Analysis of Space Exploration Vessels

Peng-Hung Tsai *,[1], Daniel Berleant *,[2]

*  Department of Information Science, University of Arkansas at Little Rock, 2801 S. University Ave.,
Little Rock, AR 72204, USA

[1] penghung.ts@gmail.com (corresponding author),  [2] jdberleant@ualr.edu

## Abstract

Forecasting technological advancement in complex domains such as space exploration presents significant challenges due to the intricate interaction of technical, economic, and policy-related factors. The field of technology forecasting has long relied on quantitative trend extrapolation techniques, such as growth curves (e.g., Moore's law) and time series models, to project technological progress. To assess the current state of these methods, we conducted an updated systematic literature review (SLR) that incorporates recent advances. This review highlights a growing trend toward machine learning-based hybrid models.

Motivated by this review, we developed a forecasting model that combines long short-term memory (LSTM) neural networks with an augmentation of Moore's law to predict spacecraft lifetimes. Operational lifetime is an important engineering characteristic of spacecraft and a potential proxy for technological progress in space exploration. Lifetimes were modeled as depending on launch date and additional predictors.

Our modeling analysis introduces a novel advance in the recently introduced Start Time End Time Integration (STETI) approach. STETI addresses a critical right censoring problem known to bias lifetime analyses: the more recent the launch dates, the shorter the lifetimes of the spacecraft that have failed and can thus contribute lifetime data. Longer-lived spacecraft are still operating and therefore do not contribute data. This systematically distorts putative lifetime versus launch date curves by biasing lifetime estimates for recent launch dates downward. STETI mitigates this distortion by interconverting between expressing lifetimes as functions of launch time and modeling them as functions of failure time. This study is the first to apply STETI within a neural network framework. Hyperparameter tuning via Bayesian optimization identified the best-performing model, which outperforms a regression-based baseline. The model's predictions across hypothetical scenarios highlight the influence of factors such as spacecraft launch mass, mission destination, and country of manufacture. The results provide insights relevant to space mission planning and policy decision-making.

## 1. Introduction

The concept of exponential technological progress has been extensively studied in technology forecasting (TF) research. Many studies have shown that various technologies follow exponential trajectories of advancement (also called Moore's law) [1–5]. This law, narrowly interpreted as the doubling of the maximum number of transistors on a microchip approximately every two years and more generally as an exponential increase in technological capability over time, has served as a useful predictor of technological progress since its discovery.

However, while many technologies exhibit trends resembling Moore's law, it remains uncertain whether this model provides the most accurate predictions. Moreover, a key limitation of relying on Moore's law for technology forecasting is its assumption that progress occurs automatically over time, without accounting for external factors. In reality, advancement is driven by continuous research, investment, and experience, not simply the passage of time [6]. Because Moore's law fails to explicitly account for these important drivers, it provides limited insight into the forces behind technological advancement. This underscores the need to move beyond time-based exponential growth models and incorporate the underlying drivers of progress to develop more accurate and insightful forecasting methods.

This study uses space exploration technology as a test case to examine a forecasting approach that integrates a modification of Moore's law to predict technological progress in this domain. Understanding the trajectory of technological progress in space exploration is critical for strategic planning and long-term investment decisions. Because space missions are becoming increasingly complex and resource-intensive, as well as economically and militarily strategic, anticipating future advancements is essential for effective resource allocation and informed decision-making. Effective forecasting helps support mission planning, research priorities, and technology policy, keeping innovation aligned with societal goals and maximizing investment returns.

While Moore's law has proven useful in predicting technological progress across various fields, its application to space exploration technology has received limited research attention, with only a few studies addressing this area [7–9]. To contribute to this research area, we propose an improvement over time-based extrapolation by integrating machine learning techniques with enhancements to the Moore's law model. Specifically, we combine Long Short-Term Memory (LSTM) networks, a specialized type of recurrent neural networks (RNNs) designed to effectively model long-term dependencies in sequential data, with an augmentation to the basic Moore's law that accounts for additional influencing factors. This multi-factor growth curve reflects expected growth patterns while accounting for a broader range of influences beyond just the passage of time. LSTMs further enhance the augmented Moore's law framework by learning complex interactions from data to uncover subtle patterns and trends that a purely time-based model could not capture. By combining the underlying structure of the modified Moore's law with the sophisticated pattern recognition capabilities of LSTMs, this hybrid approach offers a more flexible and adaptable framework for forecasting.

The aim of this hybrid approach is threefold: to improve forecasting accuracy in space exploration technology development, to identify the key drivers of progress, and to provide a decision-making tool for policymakers, researchers, and industry stakeholders. Compared to traditional time-based models, we hypothesize that this method will provide more accurate and reliable predictions by considering more factors explicitly and using machine learning to consider more factors implicitly.

The remainder of this paper is structured as follows: Section 2 offers, as background, a systematic literature review (SLR) [10] that updates our prior review of relevant studies on applying quantitative trend extrapolation techniques to technology forecasting [11]. Section 3 details the data and methods used for the research results presented herein. Section 4 presents and analyzes the results. Section 5 concludes.

## 2.      Background: Updated systematic review

The field of technology forecasting has long relied on quantitative trend extrapolation techniques, such as growth curves (e.g., Moore's law) and time series methods, to predict the trajectory of technological development. We previously conducted an SLR [11], which this review updates, to explore these techniques. It categorized key methodologies and highlighted a growing shift towards machine learning-based hybrid approaches. Building on that foundation, we now examine the current state of the field concluding that the trend is continuing and perhaps accelerating. By analyzing recent developments, this review provides an updated review that includes the latest quantitative trend extrapolation techniques in technology forecasting and informs the development of our forecasting approach for technological advancements of space exploration.

### 2.1      Implementation of the review

Our approach to this SLR implementation drew upon the process outlined by [12]. The specifics of this implementation are detailed below.

### *1. Defining research questions*

Consistent with our earlier work, we address the following key questions:

**Q1**: *What does the existing research say about the implementation of quantitative trend extrapolation methods for technology forecasting?*

**Q2**: *What quantitative trend extrapolation methods were adopted and what are the technologies where they have been applied?*

The next key question is new to this review:

**Q3**: *What does the most recent work show about the most recent trends and discoveries in this area?*

## 2. Identifying and selecting relevant studies

We followed a review protocol guided by the preferred reporting items for systematic reviews and meta-analyses (PRISMA) statement by [10] as in our previous review to develop a protocol for identifying and selecting relevant articles. Figure 1 illustrates the search strategy flowchart, followed by a description of the search and selection process. The steps involved in this process are discussed below:

• Identification

We searched Google Scholar and Web of Science for relevant publications on December 31, 2024. We tailored search queries to each database to accommodate their technical and terminological differences. We limited our search to publications from 2022 to 2024 in both databases, because 2022 was the end year of our previous review, providing continuity for this update. Additionally, we restricted results to English-language articles.

By using two queries in Google Scholar and three in Web of Science, the searches initially identified 1,143 articles: 969 from Google Scholar and an additional 174 from Web of Science. Table 1 details the final query strings and corresponding search results after a series of keyword screenings.

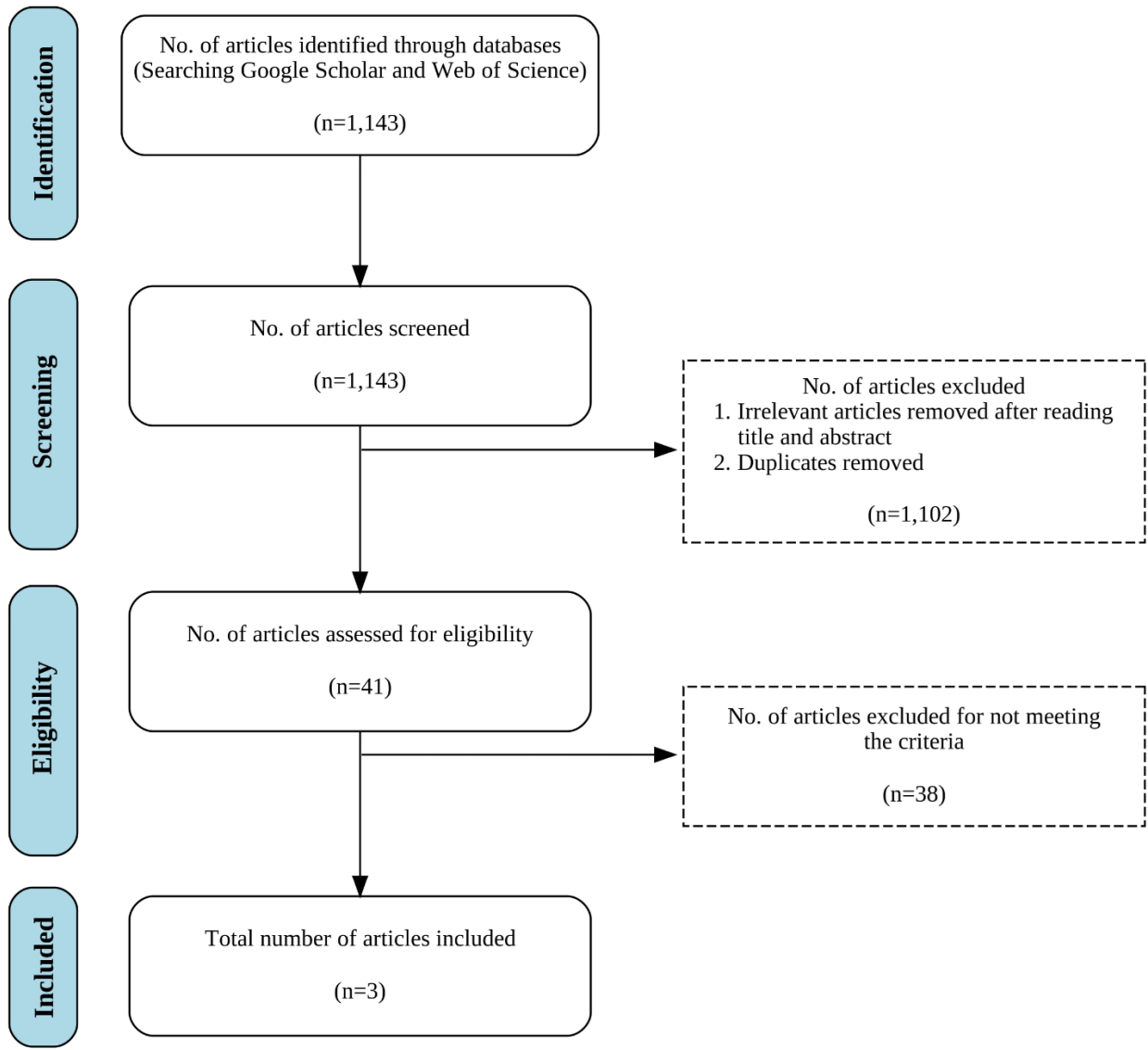

**Figure 1:** Flow diagram for the search and selection process used to develop this systematic review.

• Screening

After identifying a large pool of literature, we manually screened the titles and abstracts to narrow down the selection. We applied the following predefined inclusion and exclusion criteria to choose the most relevant articles for a thorough review.

*Inclusion criteria.* Articles had to use quantitative trend extrapolation techniques for technology forecasting, be published as journal articles, conference papers, or other types of articles (excluding theses and book chapters), and have been published no earlier than 12 October 2022, the date on which we searched both databases in our prior review.

*Exclusion criteria.* We excluded articles that focused on technology diffusion or adoption, as well as theoretical or conceptual contributions that did not use curve-fitting methods to analyze data. We also ruled out duplicates and non-English articles.

After applying these criteria, we eliminated 1,102 articles, leaving a set of 41 articles for subsequent full-text examination.

**Table 1:** Search strings and results.

| Database | Query | Search string | Results |
|---|---|---|---|
| Google Scholar | 1 | ("technology\|technological forecasting\|foresight\|prediction") ("technology\|technological progress\|progression\|advancement\|development\|evolution \|change\|growth") (quantitative\|mathematical\|statistical\|"machine learning") extrapolate error\|residual | 562 |
| | 2 | allintitle: technology\|technological\|technologies forecasting\|forecast\|forecasts | 407 |
| Web of Science | 1 | (TI=(technolog* forecast*)) AND (LA==("ENGLISH")) | 119 |
| | 2 | (TS=("technology forecasting" OR "technological forecasting" OR "technology foresight" OR "technological foresight" OR "technology prediction" OR "technological prediction")) AND TS=(quantitative OR mathematical OR statistical OR "machine learning") and English (Languages) | 37 |
| | 3 | ((ALL=("technology forecasting" OR "technological forecasting" OR "technology foresight" OR "technological foresight" OR "technology prediction" OR "technological prediction")) AND ALL=("technology progress" OR "technological progress" OR "technology progression" OR "technological progression" OR "technology advancement" OR "technological advancement" OR "technology development" OR "technological development" OR "technology evolution" OR "technological evolution" OR "technology change" OR "technological change" OR "technology growth" OR "technological growth")) AND ALL=(quantitative OR mathematical OR statistical OR "machine learning") and English (Languages) | 18 |
| Total | | | 1,143 |

- Eligibility

Next, we carefully reviewed the full texts of the remaining 41 articles to determine their eligibility. To qualify, each article had to clearly explain the forecasting techniques used, the technologies they were applied to, and the data sources and performance metrics used. Additionally, the articles had to evaluate the forecasting performance using a held-out sample and an error measure, to help demonstrate the reliability of the techniques.

After this thorough examination, 38 articles were eliminated, leaving a final set of three articles for in-depth analysis.

### 3. Interpreting the recent literature

We summarized characteristics of the selected studies to facilitate identifying trends and directions in the field as presented in subsequent sections. These characteristics are provided in Table 2.

**Table 2:** Some key properties of the articles included in this review.

| Authors | Country | Technology | Trend extrapolation methods | Data source |
|---|---|---|---|---|
| Glavackij et al. [13] | Switzerland | 148 different technologies | S-shaped curve, ARIMA, LSTM, RNN | Dataset from a previous paper |
| Dastoor et al. [14] | US | 31 technologies within NASA's 2015 Technology Roadmaps | Feedforward Neural Network (FNN) | Dataset from a previous paper |
| Zhang et al. [15] | China | Aircraft assembly technology | GRU, 2-layer-GRU, LSTM, 2-layer-LSTM, EEMD-GRU, EEMD-(2-layer-GRU), EEMD-LSTM, EEMD-(2-layer-LSTM) | The United States Patent and Trademark Office (USPTO) website database |

Next, we focus on answering the three review questions (Q1-Q3) introduced earlier.

**Q1:** *What does the existing research say about the implementation of quantitative trend extrapolation methods for technology forecasting?*

To address this question, we examined each article to highlight its contributions, methodologies, and findings.

Glavackij et al. [13] proposed a technology forecasting model based on RNNs, employing the standard RNN and more advanced LSTM architectures. They collected data from the arXiv preprint server, covering scholarly articles across 148 technical subcategories. They used the monthly number of e-prints per subcategory as a proxy for technological progress. The RNN approach was compared to S-curves (sigmoidal curves), which model growth trends that level off, and ARIMA (Autoregressive Integrated Moving Average), a statistical time-series method. The study concludes that while S-curves help explain historical trends, ARIMA is more dependable for forecasting and the RNN approach delivers superior accuracy, especially for early-stage technologies, demonstrating the potential of machine learning techniques for improving technology forecasting accuracy.

Dastoor et al. [14] developed a machine learning-based method to estimate Technology Readiness Levels (TRL), a metric for tracking technology maturity, using bibliometric data. Their analysis focused on 31 technologies drawn from the categories outlined in NASA's 2015 Technology Roadmaps [16]. The approach involved collecting data from 1995 to 2015 on scientific publications, patents, NSF awards, and articles published by NASA Spinoff via public APIs, with ground truth TRL values sourced from NASA's Technology Roadmaps. Maturity indicators were quantified by fitting annual counts of publications, NSF awards, and patents to S-curves and normalizing total Spinoff counts per technology by publication and patent totals to estimate each technology's position in its development cycle. A feedforward neural network (FNN) was then trained on these normalized S-curve values and Spinoff norms to predict a continuous "adjusted TRL," which was then broken out into individual TRLs. Technologies were then classified into Emerging (TRL1–5), Growth (TRL 6–7), and Mature (TRL 8–9) phases. Their method offers a more efficient alternative to traditional expert-based TRL assessments.

Zhang et al. [15] introduced a hybrid method, EEMD-GRU, and applied it to patent data related to aircraft assembly technology to predict the improvement trend of the technology. Ensemble Empirical Mode Decomposition (EEMD), a signal processing technique, was used to decompose the patent count time series data into multiple intrinsic mode functions (IMFs) to separate noise and extract key trend patterns. Gated Recurrent Unit (GRU) neural networks then predicted each decomposed component separately, and the final forecast was obtained by aggregating the individual predictions. The EEMD-GRU method exhibited the best predictive performance across most of the technologies, surpassing both traditional (LSTM, 2-layer-LSTM, GRU, and 2-layer-GRU) and other hybrid (EEMD-LSTM, EEMD-(2-layer-LSTM), and EEMD-(2-layer-GRU)) methods.

**Q2:** *What quantitative trend extrapolation methods were adopted and what are the technologies where they have been applied?*

To better understand the types of methods being used as well as to facilitate comparison and analysis, the methods used in the selected studies were categorized under broader method classes: (i) growth curve, (ii) machine learning, and (iii) time series, as in [11]. Table 3 provides an analysis, listing each method, its assigned class, the contributing authors, and the technologies to which it was applied. The table indicates the current prevalence of machine learning methods, particularly RNN-based algorithms such as RNN, GRU, and LSTM. Additionally, it highlights a novel application of machine learning-based hybrid methods, specifically EEMD-GRU, EEMD-LSTM, EEMD-(2-layer-GRU), and EEMD-(2-layer-LSTM), all of which combine signal processing with recurrent neural networks.

**Table 3:** Summary of the method classes and forecasted technologies used in the included studies.

| Method class | Method | Technology | Authors |
|---|---|---|---|
| Machine Learning | Feedforward Neural Network (FNN) | 31 technologies within NASA's 2015 Technology Roadmaps | Dastoor et al. [14] |
| | Gated Recurrent Unit (GRU) | Aircraft assembly technology | Zhang et al. [15] |
| | Long Short-Term Memory (LSTM) | 148 different technologies | Glavackij et al. [13] |
| | | Aircraft assembly technology | Zhang et al. [15] |
| | Standard recurrent neural network (RNN) | 148 different technologies | Glavackij et al. [13] |
| | 2-layer-GRU | Aircraft assembly technology | Zhang et al. [15] |
| | 2-layer-LSTM | Aircraft assembly technology | Zhang et al. [15] |
| | EEMD-GRU | Aircraft assembly technology | Zhang et al. [15] |
| | EEMD-LSTM | Aircraft assembly technology | Zhang et al. [15] |
| | EEMD-(2-layer-GRU) | Aircraft assembly technology | Zhang et al. [15] |
| | EEMD-(2-layer-LSTM) | Aircraft assembly technology | Zhang et al. [15] |
| Growth curve | S-shaped curve | 148 different technologies | Glavackij et al. [13] |
| Time Series | Autoregressive integrated moving averages (ARIMA) | 148 different technologies | Glavackij et al. [13] |

**Q3:** *What does the most recent work show about the most recent trends and discoveries in this area?*

We combined the results of our previous review with our current findings to provide an updated perspective on the use of forecasting methods and trends in their popularity, particularly in recent years. We observe that, as shown in Figure 2, although the previous results indicated growth curves as the most used method class from 1990 to October 2022 at 38.9% of the total, machine learning has since taken the lead at 35.3% with the inclusion of recent data, slightly surpassing growth curves, which now account for 34.1%. Time series remains in third place with 22.4%, while other uncategorized methods continue to have the smallest share at 8.2%. This aligns with the trend observed in Figure 3, which shows the evolution of method class frequency over time. The graph highlights a growing dominance for machine learning-based forecasting techniques, which can automatically learn complex patterns from data, over traditional growth curves and time series methods that rely on predefined mathematical models and assumptions.

For a detailed breakdown of the methods used in the reviewed articles, see Table 4, which lists the specific methods assigned to each class, along with the number of articles that employ these methods and their relative percentage within the total set of articles. For further explanations of these methods and their applications in technology forecasting, readers may refer to our prior review [11], which offers additional descriptions and discussions of these techniques.

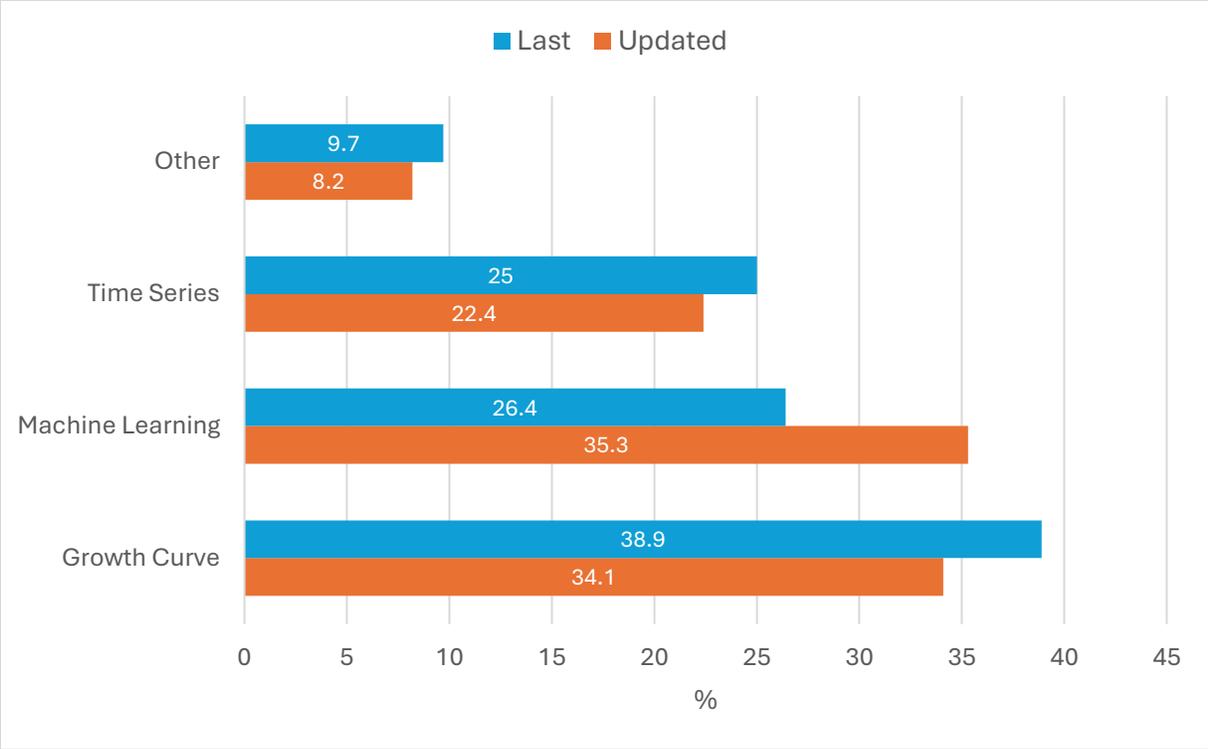

**Figure 2:** Shift in the distribution of method classes (Last review vs. Updated review). "Last" includes works published from 1990 through 12 October 2022 (the end date of the first review), while "Updated" covers works published since that date through the end of 2024.

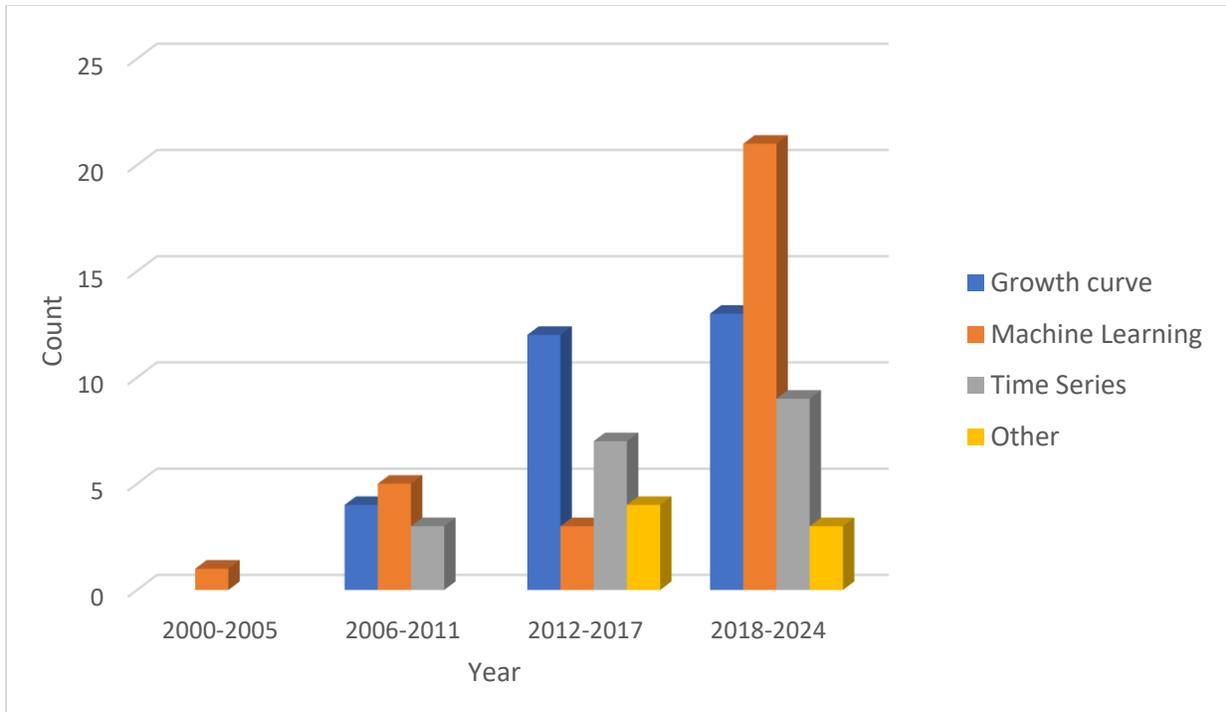

**Figure 3:** Distribution of methods, grouped by method class, across the years. Machine learning has become a key focus of the field. This is the most obvious key change in direction occurring during the 2018–2024 period.

**Table 4:** Breakdown in the methods in terms of the number of their occurrences in the included articles.

| Methods | | Article count | % |
|---|---|---|---|
| Machine Learning | | 30 | 35.3% |
| | Linear Regression | 9 | 10.6% |
| | LSTM | 6 | 7.1% |
| | GRU | 3 | 3.5% |
| | EEMD-LSTM | 3 | 3.5% |
| | SVR | 2 | 2.4% |
| | EEMD-GRU | 2 | 2.4% |
| | FNN | 1 | 1.2% |
| | LSTNet | 1 | 1.2% |
| | RNN | 1 | 1.2% |
| | Conv-RNN | 1 | 1.2% |
| | Conv-RNN-sep | 1 | 1.2% |
| Growth Curve | | 29 | 34.1% |
| | S-shaped curve | 17 | 20.0% |
| | Exponential growth law | 9 | 10.6% |
| | Experience curve | 3 | 3.5% |
| Time Series | | 19 | 22.4% |
| | ARIMA | 7 | 8.2% |
| | Exponential smoothing | 6 | 7.1% |
| | GM(1,1) grey model | 2 | 2.4% |
| | Geometric random walk with drift | 1 | 1.2% |
| | Goddard's law random walk | 1 | 1.2% |
| | Moore's law random walk | 1 | 1.2% |
| | Moore-Goddard's law random walk | 1 | 1.2% |
| Other | | 7 | 8.2% |
| | Lotka–Volterra equations | 4 | 4.7% |
| | Gupta model | 1 | 1.2% |
| | Step and wait (SAW) model | 1 | 1.2% |
| | Tobit II model | 1 | 1.2% |

## 2.2 Systematic literature review conclusion

This review of the literature indicates a growing adoption of machine learning techniques, including the use of hybrid approaches that integrate multiple methods. The reviewed studies consistently report that hybrid approaches achieved higher forecasting accuracy than single-method approaches across various technological domains. This performance advantage is likely due to the complex and nonlinear interactions among multiple factors influencing technological trends, which single-method approaches may not fully capture. By leveraging the complementary strengths of different methods, hybrid approaches provide a more comprehensive framework for handling such complexities.

With continuous advancements in machine learning and computing power, and the proven advantages of hybrid approaches, we expect to see ongoing innovation in the types of methods combined and their application to complex forecasting challenges. If these trends continue, hybrid approaches will continue to assume an increasing role in technology forecasting methodologies.

## 2.3    Discussion of the updated review

This study builds on our prior findings by confirming the growing adoption of machine learning techniques and the increasing focus on hybrid models, particularly the integration of RNN-based LSTMs and GRUs, which have been found to excel in handling sequential data.

Although the adoption of machine learning-based hybrid methods is increasing, this does not imply their absolute superiority. Methodological effectiveness depends on the characteristics of the data and the domain of application. We see opportunities for researchers in this field to evaluate these methods by conducting comparative studies. This could involve implementing hybrid models in new projects or re-assessing prior work by comparing the performance of these models against the originally used techniques.

Finally, our findings indicate that this field has evolved significantly. The rapid introduction of new techniques in recent years suggests that more new techniques will likely emerge in the near future. Thus, there will be a need for future reviews to help researchers and practitioners stay up to date with emerging advancements in technology forecasting.

In the next section, we shift attention from the existing literature to the present study.

## 3.    Methodology of the study

This methodology builds upon a key insight from the literature review: the growing role of machine learning-based hybrid methods in technology forecasting. We propose an integrated forecasting approach that combines LSTM networks with an augmentation of Moore's law. This modification incorporates additional parameters besides time that may influence technological advancement of space exploration. The LSTM learns complex interactions between these parameters, complementing the established insights provided by the Moore's law (exponential) component of the augmented model. Through this methodology, we aim to develop a more adaptable forecasting framework that enhances our ability to predict the trajectory of technological progress for space exploration vessels.

A key challenge in applying this approach is defining an appropriate metric for measuring technological progress in this domain. Unlike the computing domain, where indicators such as transistor density and processing speed clearly reflect advancement, space exploration lacks a single, directly quantifiable measure. This makes it necessary to rely on proxy variables that can capture trends in technological development.

Recent studies have proposed spacecraft lifetimes as such a proxy, suggesting that longer operational durations reflect improvements in engineering, materials, and mission design [7–9,17,18]. Inspired by this approach, we use spacecraft lifetimes as a measurable indicator whose behavior will be modeled to forecast technological advancement in space exploration.

However, applying a time-based growth curve to predict lifetimes based on launch dates runs into a fundamental problem. Because vessel lifetimes are unknown at launch and often remain unobserved for many years, only short-lived vessels contribute survival data in the near term. This leads to an overrepresentation of early failures among recent launches, introducing a form of recent-year data bias. As a result, average lifetimes for vessels launched in more recent years are systematically underestimated if not handled appropriately. This recency bias can impact lifetime analyses in numerous domains, from spacecraft operational lifetimes to engineering reliability to cancer survival and other medical domains.

This pattern of recent-year data bias is illustrated in Figure 4. Black disks represent the lifetimes of failed spacecraft plotted by their launch year as of year 2000, while black triangles show operational, non-failed spacecraft, plotted by launch year and current age. The green curve is a 15-point moving average of the observed lifetimes (black disks). It increases over earlier decades, reflecting technological improvements, but declines sharply for more recent launch years. This downward trend does not imply that newer vessels are less reliable. Instead, it results from the fact that many long-lived spacecraft launched in recent years have not yet failed and therefore do not contribute to the lifetime data.

The black curve marks the boundary of this bias, indicating the region beyond which full lifetimes are not yet available. In contrast, when the same lifetime data are plotted by failure year instead of launch year (shown as red X marks), no such downward trend is visible. This difference highlights how launch-year based lifetime estimates can be systematically biased downward in recent years if right censoring is not properly accounted for.

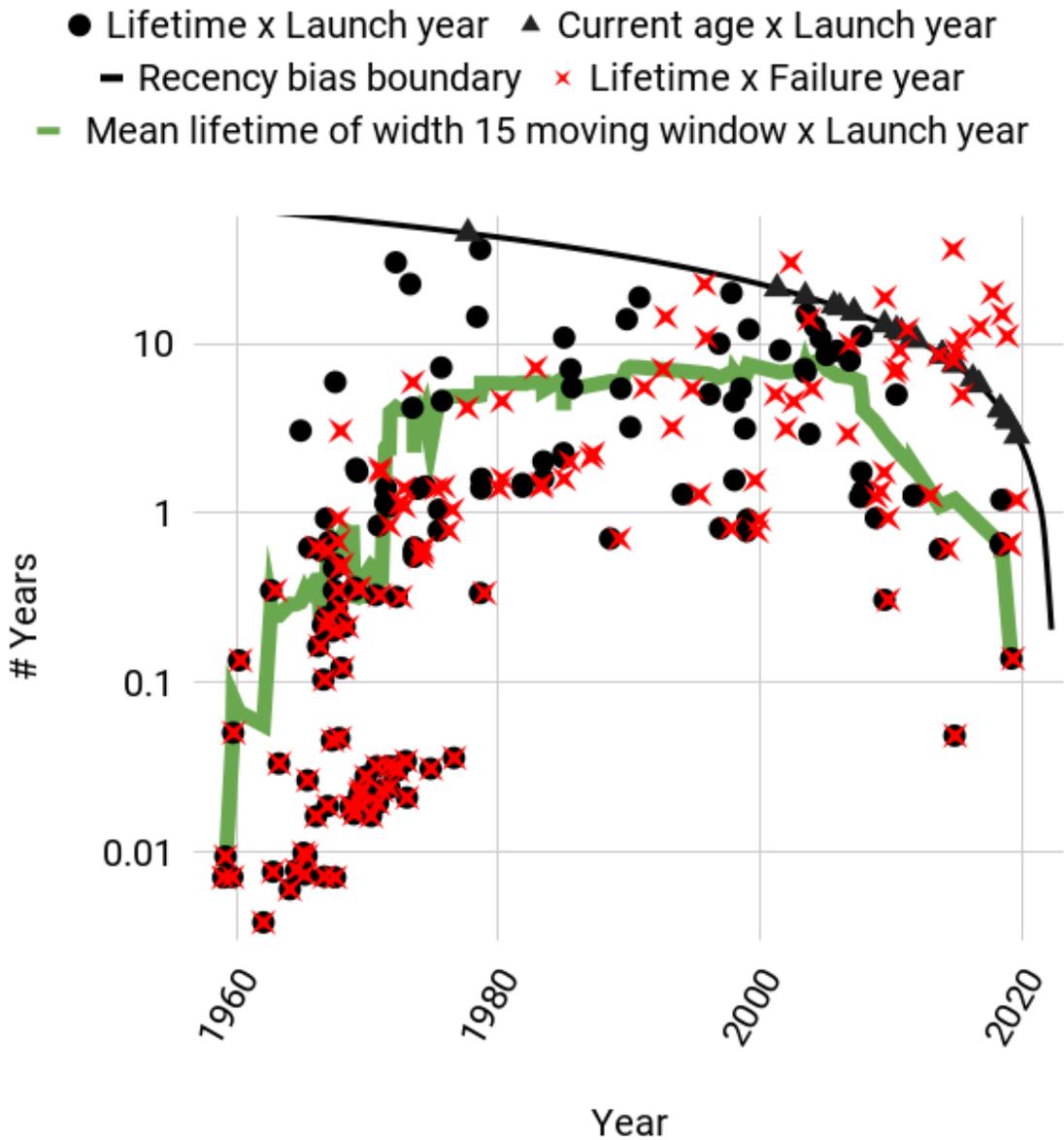

**Figure 4:** Spacecraft lifetime versus launch year. Lifetime is labeled as "#Years". The green curve, smoothed by a window size of 15 (except less where windows centered around a particular data point are limited by proximity to the end of the data sequence), shows a downturn in recent years due to recency bias from right censoring.

## 3.1 Modeling approach

To address the recent-year data bias, we propose a two-stage modeling approach termed *start time end time integration* (STETI), which enables the use of recent failure data that would otherwise either bias the results or be left unused. This method reformulates the prediction task from a launch time-based model as a failure time-based one, and then reverts the results back to the original launch time-based framework to retain the desired ability to predict lifetime from launch date. The STETI approach was introduced in [19] where it was applied to interconverting regression curves of lifetimes described as functions of start time points and lifetimes described as functions of end time points. Here we apply STETI to predict lifetimes using, instead of regression curves, artificial intelligence, specifically LSTM neural networks.

### 3.1.1 STETI fundamentals

We begin with the conceptual and mathematical foundation of STETI, which is used in our LSTM-based method. The main idea is to transform the conventional exponential equation for technological advancement over time, from modeling technological advancement as a function of date of manufacture (in general the start time, and for spacecraft specifically the launch time) into a function of the end time (for spacecraft, the date it ends operation, which for convenience we will call the failure time, although it could end operation for another reason). Thus, Eq. (1), which predicts spacecraft lifetime $l_L$ from launch time $t_L$, is transformed into Eq. (2), which predicts spacecraft lifetime $l_F$ from failure time $t_F$.

$$l_L(t_L) = l_{1959} * 2^{(t_L - 1959)/d} \qquad (1)$$

$$l_F(t_F) = l_{1959} * 2^{(t_F - l_F - 1959)/d} \qquad (2)$$

where $t_L = t_F - l_F$ represents launch time as the difference between the failure time and the lifetime. Both equations describe the lifetime value but from different perspectives: one based on launch time, $t_L$, and the other based on failure time, $t_F$. The parameters, $d$ and $l_{1959}$, respectively, represent the doubling time and the lifetime predicted by the model for spacecraft launched in 1959. The year 1959 was chosen because it marks the launch of Luna 1, the first spacecraft in our deep space exploration dataset and the first to travel beyond Earth's gravity and approach the Moon, signifying the beginning of deep space exploration. The best-fit values for $d$ and $l_{1959}$ can be found after fitting Eq. (2) to all available data on completed lifetimes, thus avoiding the aforementioned bias issue in using launch date based lifetimes. Both values, once found, are then plugged back into Eq. (1) to give the original, desired function, where launch time is the predictor variable.

### 3.1.2    LSTM-based STETI

In the LSTM-based approach to the STETI method, instead of plugging parameters fitted from the failure time based model back into the launch time based model, we transfer model outputs: predictions from the failure time model are used as training targets to train the launch time model. This mirrors the plug-back mechanism but is adapted to a deep learning setting, where parameters such as doubling time are not explicitly estimated but are learned implicitly by the neural network. The following section presents the model development procedure implemented using the LSTM-based approach to the STETI method.

### 3.2    Model development with LSTM-based STETI

Model development is carried out in two phases. The first modeling phase establishes a baseline model using time as the sole input, while the second extends this time-only baseline by incorporating additional predictors that appear to influence the trajectory of technological advancement in spacecraft, allowing us to assess the impact of this added information on predictive accuracy. Each modeling phase is implemented using the STETI two-stage modeling process, starting with failure time-based training and then transferring output predictions by using them to train the launch time-based models. Specifically, we first train a set of LSTM models on failure time-based data, to avoid them learning to mimic the misleading bias present in launch time-based data. After evaluating their performance on a held-out test set, we select the most accurate model and use its predictions as training targets to train a second set of LSTM models on launch time-based data. This output-transfer strategy enables leveraging historical failure time data, which is desired because it is free from a critical source of bias inherent in launch time data, to generate predictions for future launch times.

### 3.2.1    Phase 1: Time-only modeling

### 3.2.1.1  Phase 1, Stage 1: Failure time-based modeling

We started by training a set of LSTM models with the same architecture, varying the train/validation/test split ratios and batch sizes, to have them learn the relationship between spacecraft lifetime ($l_F$) and failure time ($t_F$), as described by Eq. (2). We first applied a logarithmic transformation to both sides of Eq. (2), resulting in the additive form shown in Eq. (3). This transformation simplifies the nonlinear relationship between variables by converting it into a linear form, which better aligns with the strengths of LSTM-based learning and improves stability during training.

$$\log_2 l_F(t_F) = \log_2 l_{1959} + (t_F - l_F - 1959)/d \qquad (3)$$

Each spacecraft in the dataset provides a data point $(t_F, l_F)$, representing its time of failure and the corresponding lifetime achieved at that point. To help capture hidden dependencies, instead of relying solely on the set of such data points to learn from, each training example fed to the LSTM consisted of, for a window size $n$, a chronological sequence of $n$ data points. For the $k$th failed spacecraft, the training example would be $X_k = (l_{F,t,k-n+1}, l_{F,t,k-n+2}, \ldots, l_{F,t,k})$, where each element in the example represents the failure time and corresponding lifetime of a spacecraft, and the window contains spacecraft $k$ and its $n-1$ preceding spacecraft failures. Moreover, the actual lifetime of the spacecraft was used as the training target. Using these input–target pairs, the LSTM is trained to learn the mapping from a sequence of failure time–lifetime pairs to the lifetime of a spacecraft. The LSTM produces a lifetime prediction based on the input sequence $X_k$. This is represented by Eq. (4):

$$\log_2 l_{F,k} = f_1(X_k; \theta_F) \qquad (4)$$

where $f_1(.)$ is the function learned by the LSTM, $\theta_F$ denotes LSTM weights and biases, $\log_2 l_{F,k}$ is the log-transformed predicted lifetime of spacecraft $k$, consistent with Eq. (3).

Once training of the models was complete, we evaluated their predictive performance on the test data and selected the one with the best results. Its predicted lifetimes were then used as training targets to train another set of LSTM models in Stage 2.

### 3.2.1.2  Phase 1, Stage 2: Launch time-based modeling

Similar to the transformation applied to the failure time model in Stage 1, we apply a logarithmic transformation to the launch time model described by Eq. (1). This results in the additive form shown in Eq. (5).

$$\log_2 l_L(t_L) = \log_2 l_{1959} + (t_L - 1959)/d \qquad (5)$$

We next trained a second set of identically structured LSTM models, again varying the data splits and batch sizes. To better capture hidden dependencies, we did not rely solely on individual data points. Instead, each training example fed to the LSTM consisted of a chronological sequence of $n$ launch times, corresponding to a sliding window of size $n$. For the $k$th launched spacecraft, the training input would be $Y_k = (t_{L,k-n+1}, t_{L,k-n+2}, \ldots, t_{L,k})$, where each element represents the launch time of a spacecraft, and the window contains spacecraft $k$ and its $n$-$1$ preceding spacecraft launches. Additionally, following our output-transfer strategy, we used the output from the best-performing model in Stage 1 as the training target. Using these input–target pairs, the LSTM is trained to learn the mapping from a sequence of launch times to a predicted spacecraft lifetime. The LSTM produces a lifetime prediction based on the input sequence $Y_k$. This is given by Eq. (6):

$$\log_2 l_{L,k} = f_2(Y_k; \theta_L) \qquad (6)$$

Here, $f_2(.)$ is the LSTM-learned function, $\theta_L$ represents the learned weights and biases, and the model outputs $\log_2 l_{L,k}$, a log-transformed lifetime prediction.

After training all models, we evaluated their predictive performance on the held-out test data. The model achieving the highest predictive accuracy was selected as the best model.

### 3.2.1.3 Recap of the elements of the method

1. The objective is to predict the expected lifetime of future spacecraft based on their launch dates, leveraging patterns learned from past spacecraft lifetimes.

2. This task is impeded by the fact that lifetimes of recently launched spacecraft are either relatively short, or are unknown because the spacecraft have not yet failed. This censoring problem biases the trajectory of lifetime expectations downward even if lifetimes are actually increasing.

3. To circumvent this censoring bias, we shift from analyzing data on lifetimes associated with launch dates to analyzing data on lifetimes associated with failure dates.

4. An LSTM model is developed to predict spacecraft lifetimes given their failure dates. This intermediate step mitigates recent-year data censoring bias (Fig. 4) and thus supports the next step, forecasting of future spacecraft lifetimes based not on failure dates but on launch dates as desired.

5. The LSTM predictions of lifetimes and associated launch dates are used as training data for another LSTM that will then be able to predict a lifetime for any recent or hypothetical future launch date.

This process is extended to more complex models of spacecraft lifetime next.

### 3.2.2 Phase 2: Time-plus modeling

In the second modeling phase, we expand the time-only models by incorporating other potential predictors of technological change in space exploration, aiming to assess their impact on lifetime prediction performance.

In our prior work [9], we introduced a novel approach that incorporated an additional variable, the NASA yearly budget, into a generalized Moore's law formula, considering the indispensable role that continued government funding plays in NASA funded research and missions. The study found that the budget-augmented model achieved greater accuracy in fitting the data than the general exponential curve. The results also showed that the NASA budget has had a statistically significant impact on spacecraft lifespans, highlighting the importance of continued government funding in boosting technical performance of future spacecraft. Thus, we incorporate NASA budgets in this study.

In search of other potentially relevant predictors, we chose to include non-technical features, as technical features can become less relevant in predicting the future when a given technical approach that offers those technical features is replaced by a successor approach offering new features [20]. For example, while screen resolution was the main competing element for earlier CRT monitors, compactness became the key dimension for current LCD monitors [21].

Many studies, such as [5,21–25], have emphasized the indispensable role of R&D investments in technological development or suggested using R&D investments as an input for technology forecasting tasks. Adding this variable, therefore, would likely be useful for enhancing the model forecasting power.

Additionally, we experimented with a spacecraft-specific feature, spacecraft launch mass, which is the total mass of spacecraft (including the propellant) before launch, to examine how it might affect the models' prediction ability. Mission properties such as the type of contact (orbit, hard landing, soft landing, rover, etc.) and destination (moon,

Venus, etc.) were considered as well to assess how they might predict lifetimes.

We also investigated whether and how the evolution of spacecraft lifetimes varies across countries operating the spacecraft, as different countries possess different levels of space capabilities. While some countries have a well-developed domestic space industry and are able to generate their own space technologies on their own, others are only capable of developing a part of their own space technologies and have to partner with more advanced space players to achieve what they could not do alone. Howell et al. [8] compared the changes in spacecraft lifespans over time between the United States and the Soviet Union during the space race and found that the latter had longer average lifetimes at the beginning but the former made faster progress, gaining the lead later. Our study further looks into how spacecraft lifetimes evolve differently for different countries.

### 3.2.2.1  Phase 2, Stage 1: Failure time-based modeling

Using the two-stage process described above, we began by training a set of models with the same LSTM architecture on failure times, lifetimes, and the newly introduced predictors, using various data splits and batch sizes. Eq. (7), an extension of Eq. (4), defines the model form:

$$\log_2 l_{F,k} = h_1\big(X_k, U_{F,t-n+1:t}, V_{F,k}, M_{F,k}; \theta_F\big) \qquad (7)$$

where $h_1(.)$ denotes the function learned by the LSTM such that $\log_2 l_{F,k}$ represents the predicted log-lifetime of the failed spacecraft $k$. The sequence $U_{F,t-n+1:t}$ represents a backward-looking window of $n$ consecutive yearly values of R&D investments and NASA budgets, ending in the year corresponding to the failure of spacecraft $k$ at time $t$. It includes data from $U_{F,t-n+1}$ to $U_{F,t}$. $V_{F,k}$ captures the categorical features of spacecraft $k$, which failed at time $t$, including the country factor and mission characteristics such as contact type and destination. The spacecraft launch mass is denoted by $M_{F,k}$, while $\theta_F$ represents the internal weights and biases learned by the LSTM. Eq. (7) is basically Eq. (4) with extra features added, which leads to a different set of parameters $\theta_F$. After training was completed for the models, we compared their performance on the test data and identified the best-performing one. The predicted lifetimes from this model served as training targets for training the next set of LSTM models in the next stage.

### 3.2.2.2 Phase 2, Stage 2: Launch time-based modeling

Subsequently, we trained a new set of LSTM models, identical in architecture, using launch times and the new predictors as input features, with the predicted lifetimes from the best previous-stage model serving as targets. As described in Eq. (8),

$$\log_2 l_{L,k} = h_2\big(Y_k, U_{L,t-n+1:t}, V_{L,k}, M_{L,k}; \theta_L\big) \tag{8}$$

which extends Eq. (6), each model was trained to learn the function $h_2(.)$ by processing $Y_K$ (chronological sequences of spacecraft launch times), $U_{L,t-n+1:t}$ (sequences of yearly R&D investments and NASA budgets), the categorical features $V_{L,k}$ (country factor, contact type, and destination) at time $t$, the spacecraft launch mass $M_{L,k}$, and LSTM-specific weight and bias parameters $\theta_L$ in order to predict $\log_2 l_{L,k}$. After training was complete, the most accurate model was designated as the best model.

### 3.3 Modeling framework

A schematic of the modeling framework is presented in Figure 5. This framework outlines the key stages involved in developing and implementing the model. Each stage is represented by a distinct block, illustrating the sequential flow of processes.

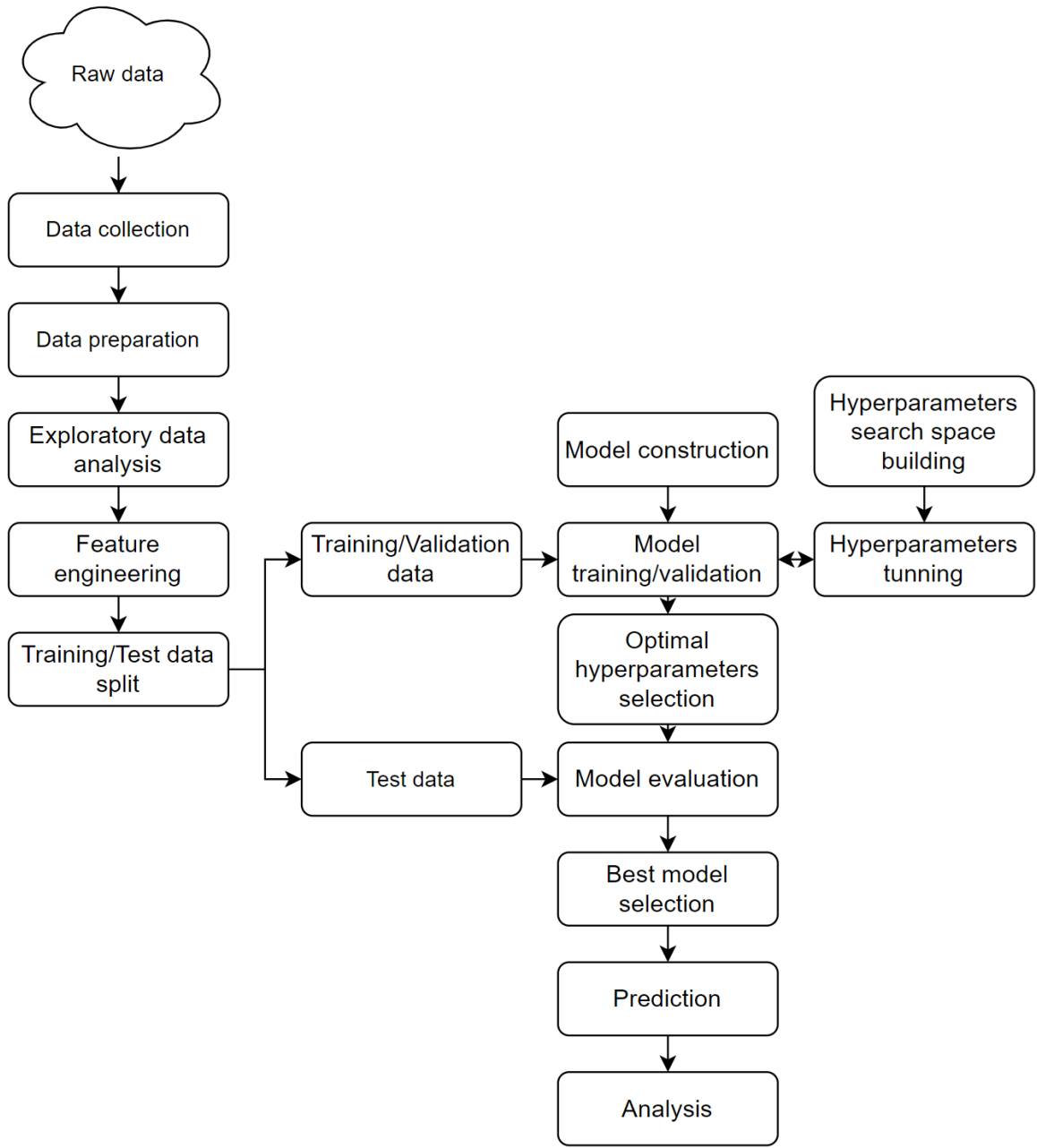

**Figure 5:** Modeling framework.

## 3.4    Data

In this section, we describe the types of space missions we investigated, and then delve into the data collection process, explaining how we gathered relevant information. We then describe the data augmentation and preparation procedures that we used to enhance the dataset's quality and completeness so that it is well-prepared for subsequent analysis. After that, we provide an overview of the dataset and give visualizations showing the dataset's key features and various trends and other patterns in the data. We end by discussing the feature engineering methods that were used to make the data more suitable for modeling.

### 3.4.1    Spacecraft for exploration of extraterrestrial objects

The spacecraft examined in this research were those with a mission objective including the exploration of at least one extraterrestrial body, excluding the Sun [7]. These vessels were designed to venture beyond Earth's immediate vicinity to explore celestial bodies and go near or to objects beyond Earth to conduct scientific investigations and advance our understanding of them. Spacecraft with exploration-driven missions are equipped with specialized instruments and technologies for interplanetary travel, deep space navigation, and scientific research in outer space environments. In this context, Earth-observing satellites are intentionally excluded from consideration, as they are designed to orbit our planet to monitor and investigate its environment, primarily benefiting terrestrial studies and interests. Their mission objectives are distinct from the broader exploration of other celestial bodies from Earth's moon and beyond. By focusing on spacecraft with such exploration missions, we aim to provide insights and analyses that enhance our understanding of the challenges and goals of deep space exploration.

### 3.4.2    Data collection

#### 3.4.2.1  Space exploration missions

The dataset utilized for analysis was expanded and refined from the work of [26]. The original dataset, collected from NASA-affiliated websites and Wikipedia up to the year 2017, contains information regarding the name of NASA-associated missions, the extraterrestrial object explored, and the type of contact made. To compile a comprehensive and up-to-date list of space missions launched globally until 2022, we conducted extensive online research, primarily relying on two sources: Wikipedia, a publicly accessible knowledge repository, and several NASA-affiliated websites, including the NASA Jet Propulsion Laboratory, NASA Solar System Exploration, and NASA Space Science Data Coordinated Archive (NSSDCA), which are reliable repositories of official information.

We began by identifying a list of relevant missions and extracting all pertinent spacecraft data from Wikipedia, including mission or spacecraft names, launch dates, failure dates (if applicable), launch masses, extraterrestrial destinations, and countries of manufacture for spacecraft. This initial source provided extensive information on missions and spacecraft. However, to ensure the accuracy and authenticity of the data retrieved from Wikipedia, we verified it by checking the original sources cited next to the data, when those sources were available. Additionally, data from non-NASA-affiliated websites was cross-referenced with NASA-affiliated sources for further validation. In cases of missing or incomplete data, or reference sources that were absent or unverifiable through NASA-affiliated websites, we cross-referenced multiple trusted and authoritative sources, such as government reports, press releases, and websites of spacecraft manufacturers or space agencies of other countries if the spacecraft was launched or developed by non-U.S. organizations, to supplement and validate the information.

### 3.4.2.2  NASA budget

Collection of NASA's annual budget was from Wikipedia [27], which also lists the 1915-1959 budget data of the National Advisory Committee on Aeronautics (NACA), the precursor to NASA. We used the 2022 Constant Dollars budget data, which adjusts for inflation and hence provides a more meaningful assessment than the Nominal Dollars budget data. We compared the data to the official websites of NASA's History Division and the Statistical Abstracts of the United States, maintained by the U.S. Census Bureau, to ensure accurate and complete data.

### 3.4.2.3  R&D investments

We obtained the federal R&D investment data from the website of the National Center for Science and Engineering Statistics (NCSES), a federal statistical agency under the National Science Foundation. NCSES maintains an online database that contains a wide range of expenditure and funding data related to R&D activities. Specifically, we focused on the following categories: Total for All Budget Functions, National Defense, and Space Flight, Research, and Supporting Activities, which were sourced from the table titled "Federal funding for R&D, R&D plant, and basic research, by budget function: FYs 1955–2023" [28]. The table presents the data in current US dollars (millions of dollars) annually, spanning from 1955 to 2023. To adjust for inflation, we converted the dollar amounts of these three categories to 2021 US dollars using the Consumer Price Index retroactive series derived from the Bureau of Labor Statistics [29].

### 3.4.3 Data augmentation & preparation

We integrated the R&D investment data and NASA budget data into the mission dataset. Subsequently, a data augmentation and preparation process was undertaken to enhance the dataset's comprehensiveness and usability. This involved standardizing mission names, transforming the date formats (launch dates and failure dates) from year-month-day to year with a decimal, and converting the attributes to the correct data types. Additionally, two new attributes were introduced: "Lifetime," calculated as the duration between a mission's launch and failure for spacecraft that experienced failure, and "Status," which indicates whether the spacecraft is operational or not. Ultimately, the dataset comprises 177 records, each representing an individual mission and covering multiple data types.

### 3.4.4 Data description & visualization

This section presents an exploratory data analysis (EDA) to gain a better understanding of the data. We begin by examining essential information about the attributes in our dataset. Table 5 provides a detailed overview of attribute descriptions, data types, and value ranges or distinct values for each attribute. As can be seen, the dataset comprises 12 attributes: "Launch date," "Failure date," "Lifetime," "Status," "Launch mass," "Destination," "Type of contact," "Country of MFR," "Total R&D," "National Defense R&D," "Space R&D," and "NASA Budget," with the first attribute "Launch date" spanning from 1959.005 for the first entry to 2022.945 for the last. The attributes are divided into two data types: "float64," which is utilized for storing numerical values with decimal points, and "object," which is used to contain text-based values (strings) in our dataset.

**Table 5:** Attribute descriptions and data characteristics.

| Attribute | Description | Data type | Values |
|---|---|---|---|
| Launch date | The specific date on which the spacecraft is launched into space. | float64 | 1959.005 - 2022.945 |
| Failure date | The specific date when the spacecraft loses the capacity to perform its objectives. | float64 | 1959.012 - 2022.956 |
| Lifetime | The duration of the spacecraft's operational life. | float64 | 0.004 - 36.094 |
| Status | Indicating whether the spacecraft is currently operational or failed. | object | ['inactive' 'active'] |
| Launch mass | The weight of the spacecraft at the time of its launch, measured in kilograms. | float64 | 6.1 - 46801.0 |
| Destination | The intended target or position in space where the spacecraft is designed to reach. | object | ['Lunar' 'Venus' 'Mars' 'Jupiter' 'Saturn' 'Mercury' 'Neptune' 'Solar' 'Comet' 'Phobos' 'Asteroid' 'Titan' 'Kuiper Belt Object' 'Asteroid belt' 'Jupiter Trojan asteroids'] |
| Type of contact | The specific manner in which a spacecraft interacts with its destination. | object | ['Impact' 'Flyby' 'Unsuccessful landing' 'Soft landing' 'Orbit' 'Orbit/Unsuccessful landing' 'Lander'] |
| Country of MFR | The country responsible for building the spacecraft. | object | ['Soviet Union' 'US' 'Japan' 'Europe' 'Europe_US' 'China' 'India' 'Luxembourg' 'Europe_Japan' 'Israel' 'UAE_US' 'South Korea_US'] |
| Total R&D | The total annual U.S. federal research and development funding, measured in USD millions. | float64 | 54619.75 - 197486.01 |
| National Defense R&D | The annual U.S. federal research and development funding for national defense, measured in USD millions. | float64 | 44440.98 - 107808.48 |
| Space R&D | The annual U.S. federal research and development funding for space flight, research, and supporting activities, measured in USD millions. | float64 | 0.0 - 36860.36 |
| NASA Budget | The U.S. federal funding allocated to NASA each year to support its various missions and initiatives in space and aeronautics, measured in USD millions. | float64 | 3536.8 - 49551.0 |

Figure 6 presents the status distribution across the spacecraft. It can be seen that of the 177 spacecraft in the dataset, approximately 80% (140 spacecraft) have failed, compared to 20.9% (37 spacecraft) that remain operational. Figure 7 shows the distribution of spacecraft lifetimes. Lifespans range widely, from zero to over 35 years. The distribution is right-skewed, with a significant concentration of lifespans between 0 and 5 years.

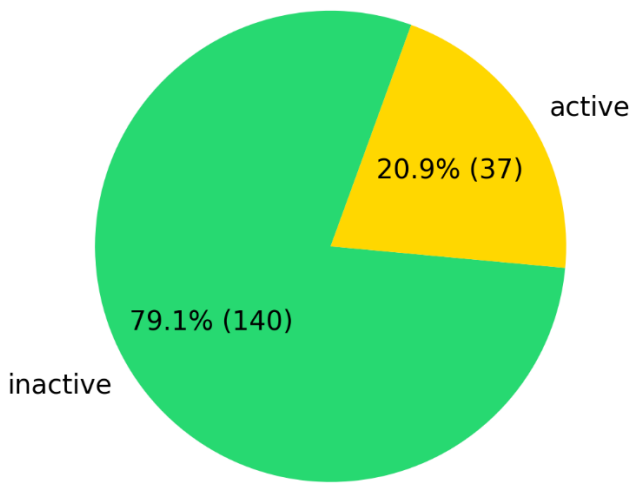

**Figure 6:** Percentage distribution of spacecraft by "Status."

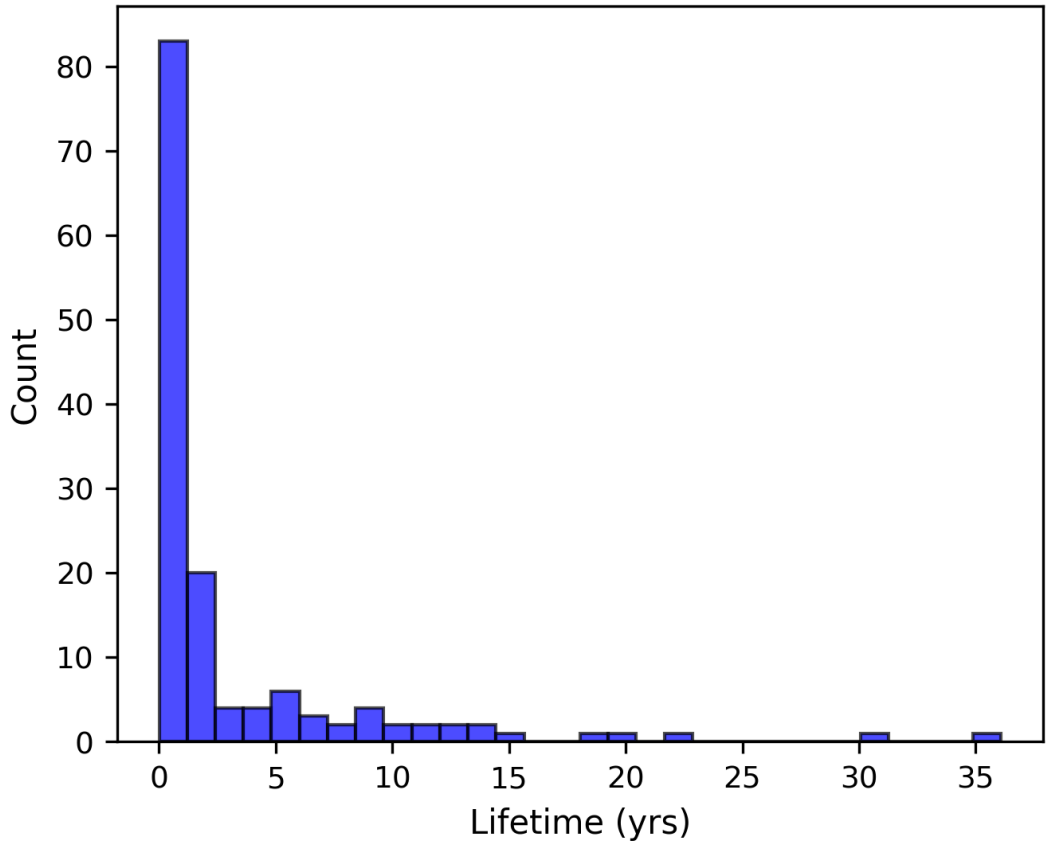

**Figure 7:** Lifetime distribution.

Figure 8 displays the distribution of spacecraft lifespans in different lifetime ranges, categorized by the country of manufacture of the spacecraft. Evidently, the United States leads other countries in the number of spacecraft launched, followed by the Soviet Union and Japan. While the United States, Europe, Japan, and India have had spacecraft that lasted longer than 5 years, only the United States has spacecraft that endured for over two decades.

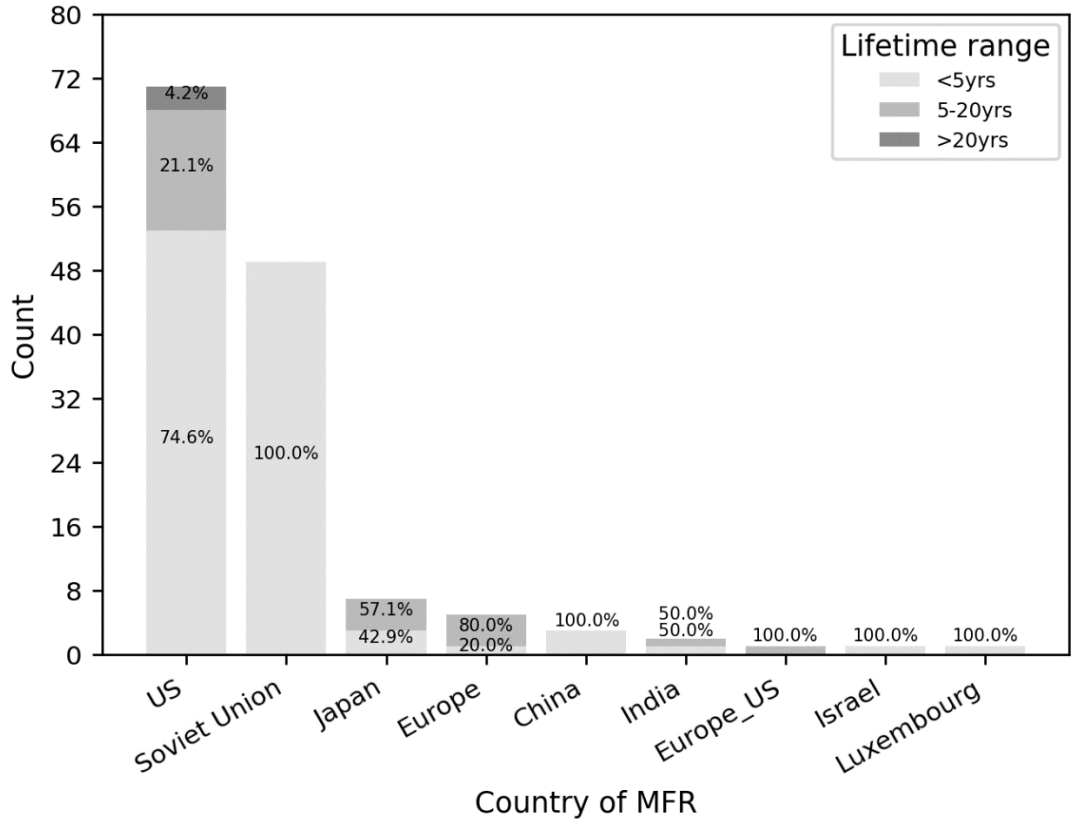

**Figure 8:** Distribution of lifetimes in different ranges by "Country of MFR," the main country of manufacture of the spacecraft (components may have been made elsewhere)

Figure 9 displays the distribution of lifespans in different lifetime ranges, grouped by the type of contact made with the destination. One can observe that the "orbit" type of contact has a higher proportion of spacecraft that lasted 5 years or longer than other contact types, and both "orbit" and "flyby" are the only contact types that have spacecraft with a lifespan exceeding 20 years.

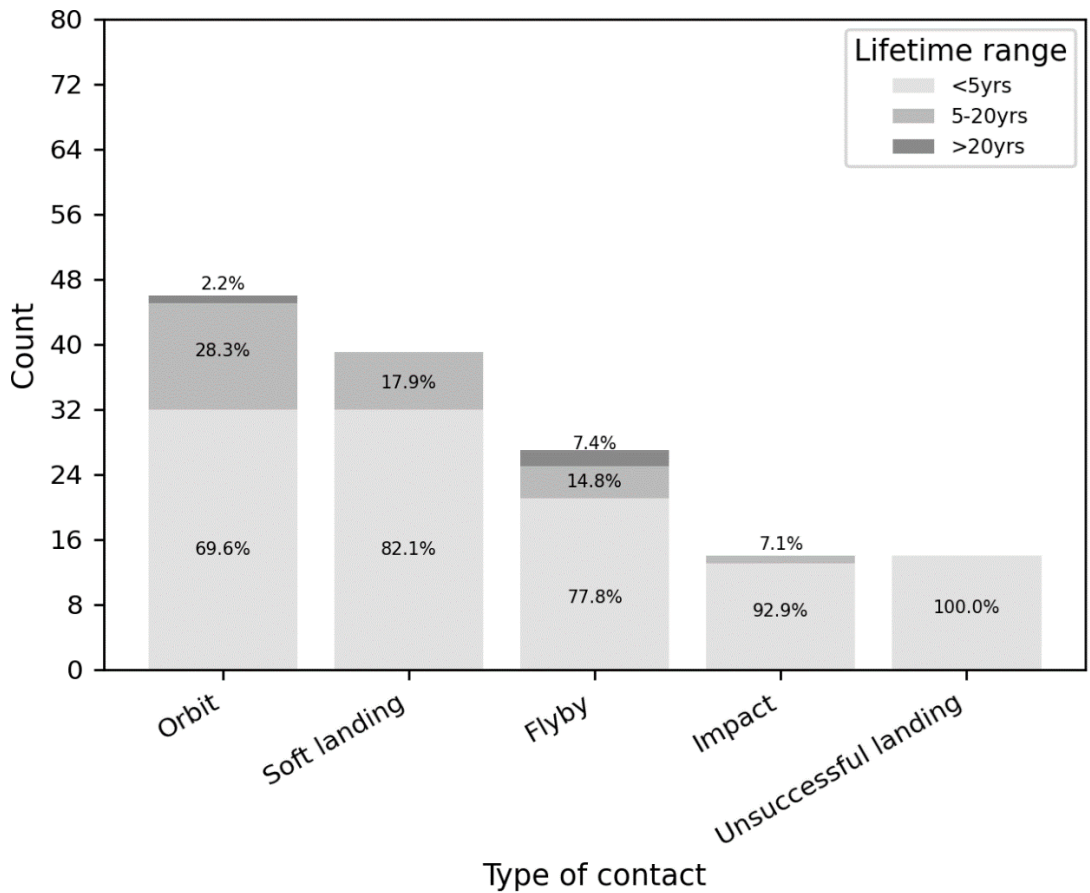

**Figure 9:** Distribution of lifetimes in different ranges by "Type of contact" with the destination body.

The distribution of spacecraft lifespans also differs across different extraterrestrial destinations, as evidenced in Figure 10. While a substantial number of spacecraft were sent to the Moon, the majority of them had lifespans of less than five years. On the other hand, spacecraft sent to destinations such as "Jupiter," "Saturn," and "Solar," which includes those operating in heliocentric or solar-adjacent orbits for other scientific purposes, were able to last for more than 20 years.

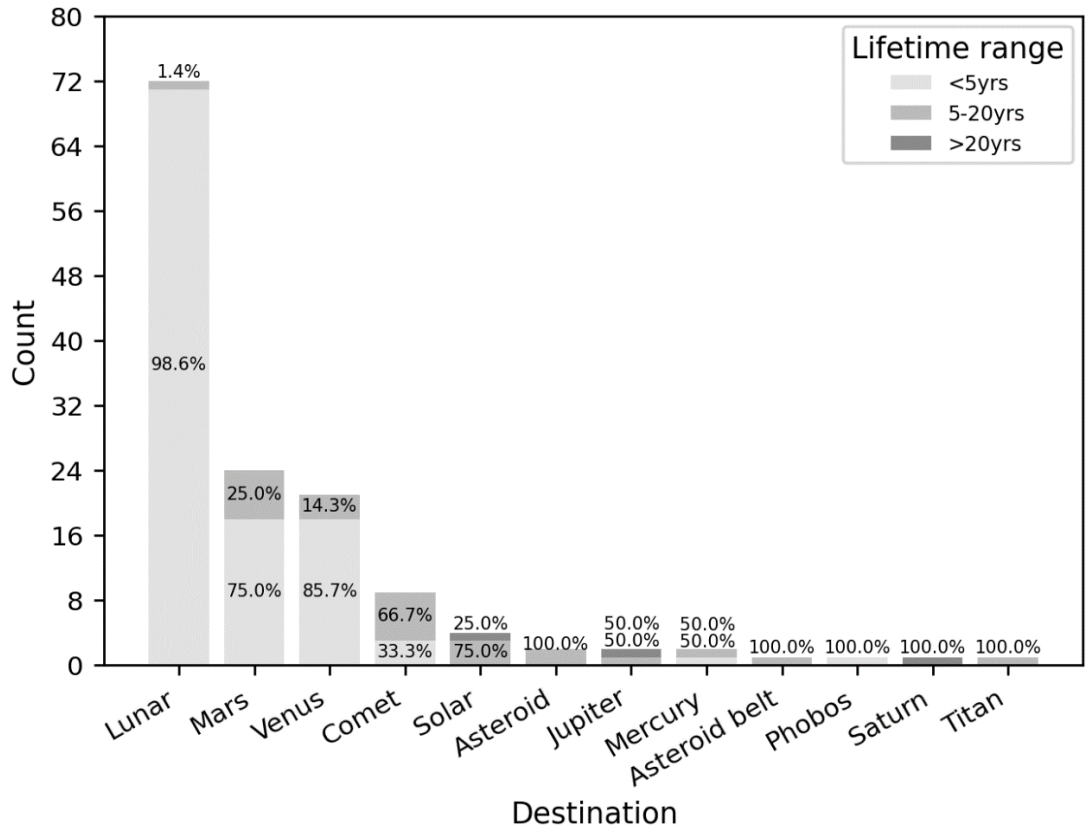

**Figure 10:** Distribution of lifetimes in different ranges by "Destination."

Figure 11 shows the relationship between spacecraft lifespan and spacecraft launch mass. Each data point on the plot represents an individual spacecraft. As can be seen, the great majority are concentrated in the lower-left corner with values under 7,500 kg.

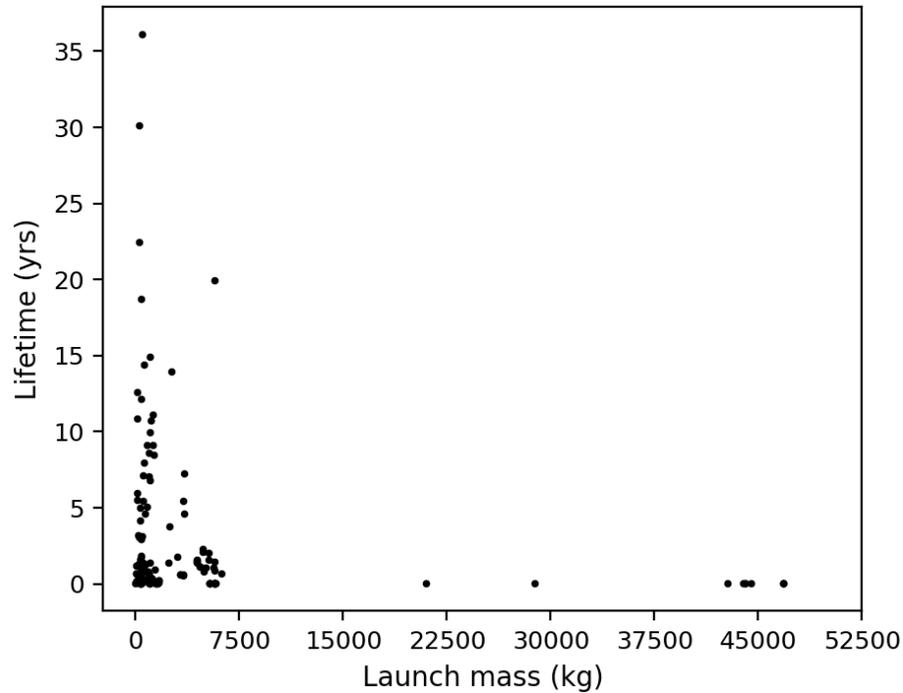

**Figure 11:** "Lifetime" vs. "Launch mass." Launch mass includes fuel.

The scatter plots depicted in Figures 12 through 15 show the relationship between spacecraft lifespan and the four U.S. federal funding attributes: "Total R&D," "National Defense R&D," "Space R&D," and "NASA Budget," respectively. The red regression trend line illustrates a linear association between the two variables, although the low R-squared values associated with these linear curves indicate less than a strong fit to the data.

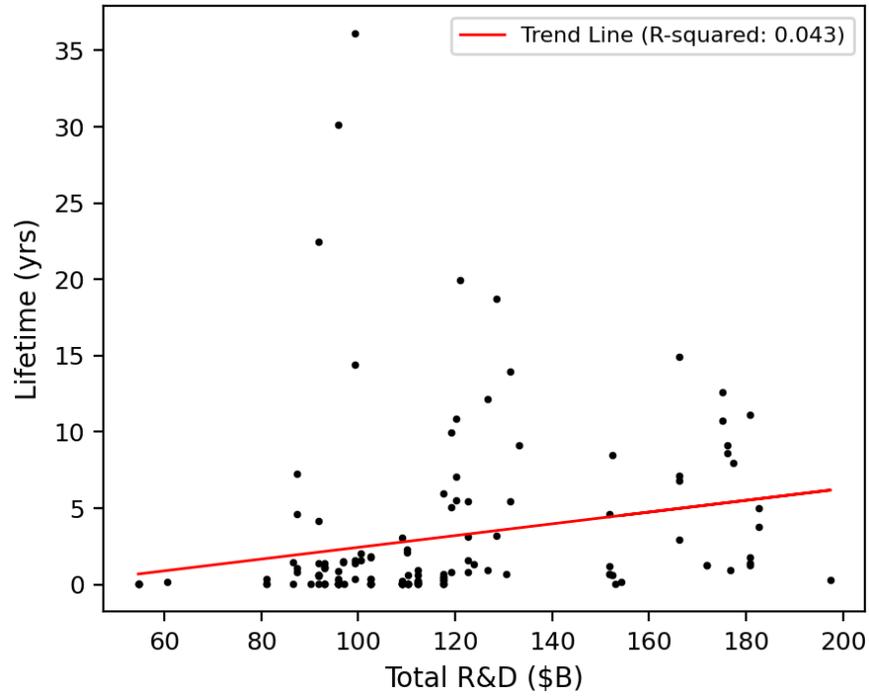

**Figure 12:** "Lifetime" vs. "Total R&D."

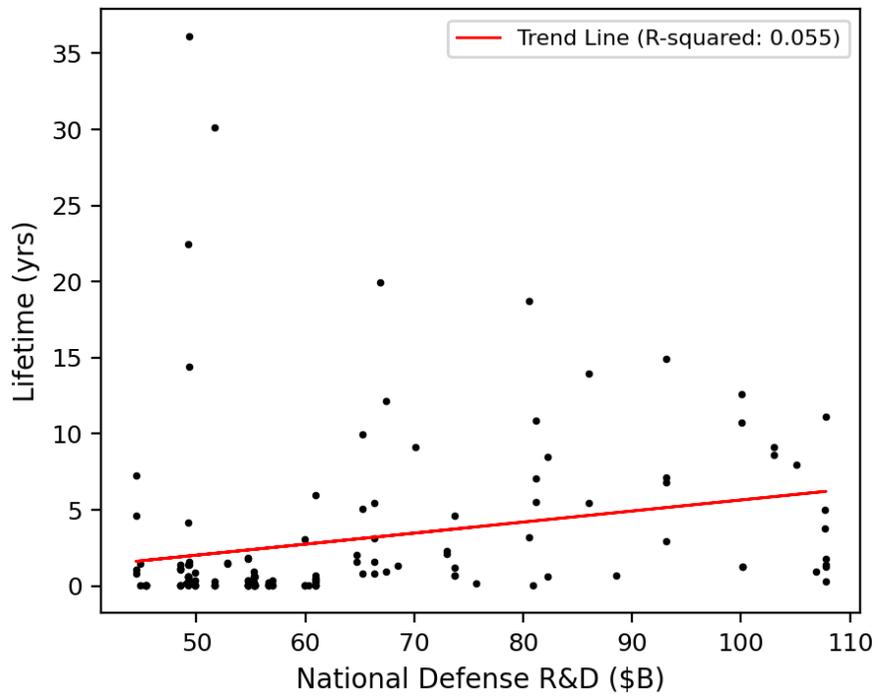

**Figure 13:** "Lifetime" vs. "National Defense R&D."

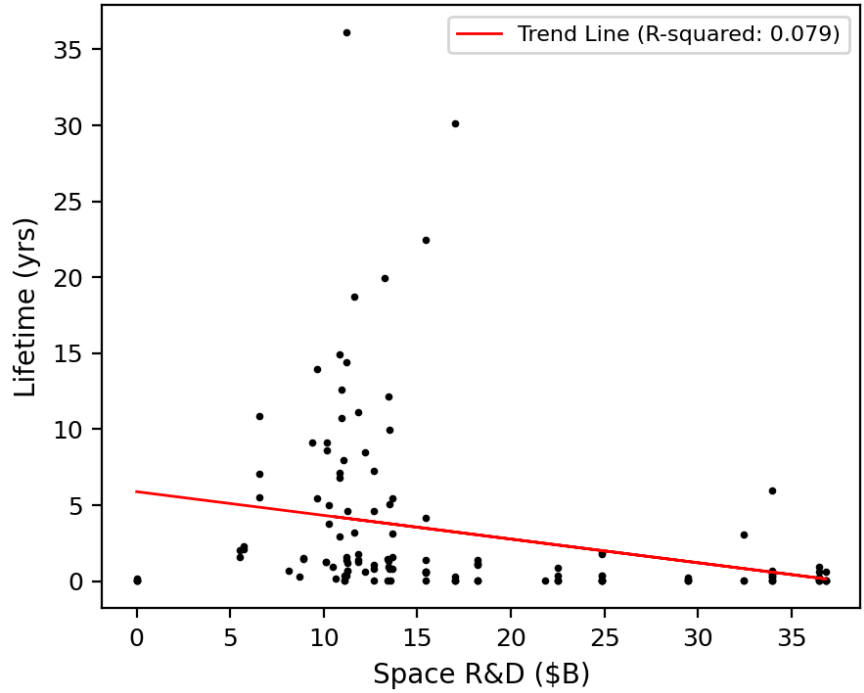

**Figure 14:** "Lifetime" vs. "Space R&D."

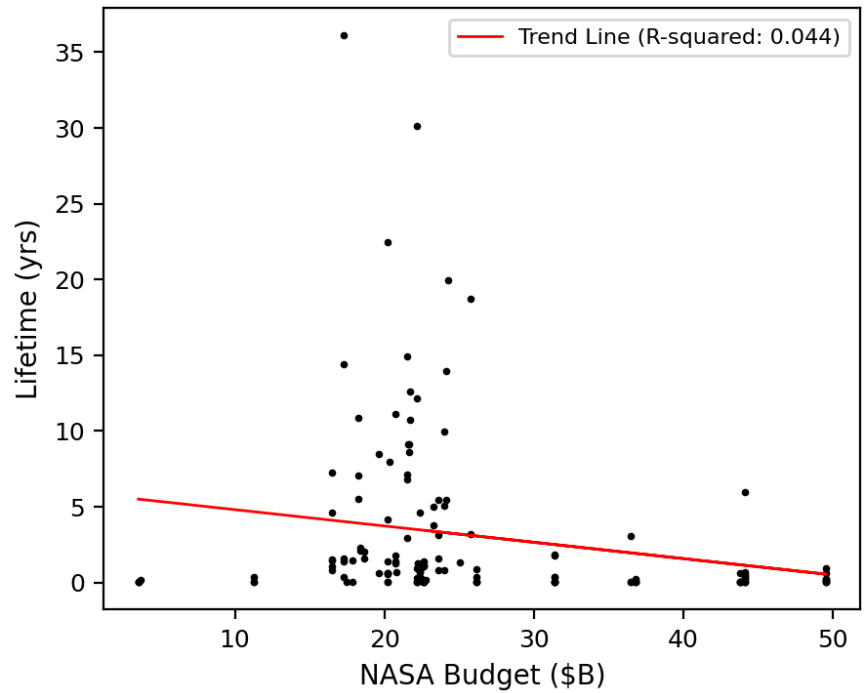

**Figure 15:** "Lifetime" vs. "NASA Budget."

This EDA provides a view of the dataset that includes the distribution of the attributes, the relationships between them, and potential areas for feature engineering. This knowledge also supports tailoring the dataset for subsequent modeling efforts.

### 3.4.5 Feature engineering

Feature engineering is the process of selecting, preparing, and constructing features that are better suited for machine learning algorithms [30]. It can involve identifying new features, handling missing data, encoding categorical features, and scaling the data. Prior studies have shown that feature engineering can enhance the efficiency and accuracy of machine learning models [31]. While feature engineering is a critical part of machine learning and can have a significant impact on model performance, there is no universally applicable approach. The preferred method is often through experimentation with different techniques on the data to observe their impact on model performance. Furthermore, in the context of neural networks, feature engineering is of lesser concern, as neural networks can automatically learn and discover meaningful features directly from raw data during the training process. Nevertheless, we still apply deliberate feature transformations with the aim of enhancing the model's overall performance.

### 3.4.5.1 Feature scaling

We rescaled the numerical features in our dataset to the range 0 to 1 to prevent features with larger ranges from dominating the others during the training process.

### 3.4.5.2 Logarithmic transformation

For modeling purposes, we performed a logarithmic transformation on the "Lifetime" attribute to create the target variable, "log2(Lifetime)." This reduces its skew, making it more normally distributed, which was expected to improve model performance.

### 3.4.5.3 Sliding window

Our numerical data comprises features such as "Launch date," "Failure date," "Lifetime," "Total R&D," "National Defense R&D," "Space R&D," "NASA Budget," and "Launch mass." The features were structured into a sliding window format, which formed a lag or window of consecutive data points. Each window formed an input example for LSTM training. The window slid through the data, providing the LSTM with a sequence of input data and their corresponding output or prediction targets. The best window size or sequence length was determined through hyperparameter optimization during the training process. Figure 16 illustrates how the sliding window method was applied to the data. In the figure, each number (1, 2, 3, ..., 9) represents an observation at a specific time or step in the data. In the initial window, the first five observations (from steps 1 to 5) are included within the window and used to make a prediction about the current target or output. After that, the window slides one step to the right. This means that it now uses a new set of observations, ranging from steps 2 to 6, to predict the next target. The process continues, one step at a time until all the data is processed.

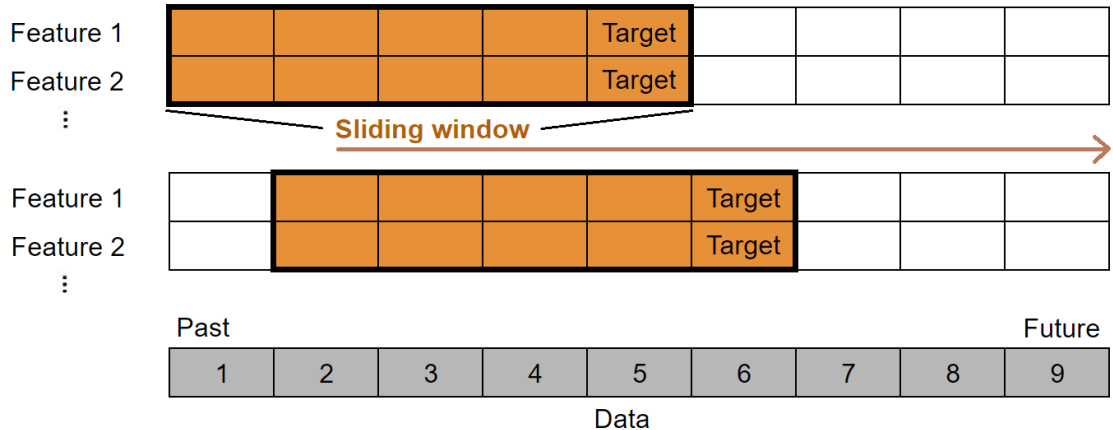

**Figure 16:** Sliding window process. Each window constitutes a training instance.

## 3.5 Theoretical background

We now review the foundational LSTM and Bidirectional LSTM neural networks that constitute our model architecture. We also introduce the optimization technique employed for hyperparameter tuning and outline the data splitting method used. Furthermore, we detail the performance metrics utilized in evaluating the models.

### 3.5.1 LSTMs

Long Short-Term Memory networks (LSTMs), introduced by [32], are a type of recurrent neural networks (RNNs) that are effective at learning long-range dependencies in time series or sequential data. Standard RNNs rely on simple recurrent units that use the internal hidden state alone to remember any length of input data and are prone to short-term memory problems due to vanishing or exploding gradients. LSTMs overcome this challenge by incorporating memory cells and controlling gates to store and update information over longer sequences. Unlike feedforward neural networks, which process inputs independently, LSTMs process inputs sequentially, enabling them to learn relationships between data points in a sequence. LSTMs have been widely investigated for sequential data tasks such as technology forecasting, speech recognition, and machine translation [33]. They have also achieved success in multivariate time series prediction owing to their ability to learn complex data features over extended temporal contexts [34,35]. Spacecraft lifetime data exhibits a temporal correlation given that the current state of spacecraft technology significantly influences the near-term future state. Establishing an LSTM-based prediction model for spacecraft lifetimes aligns with that characteristic, suggesting the potential value of doing so.

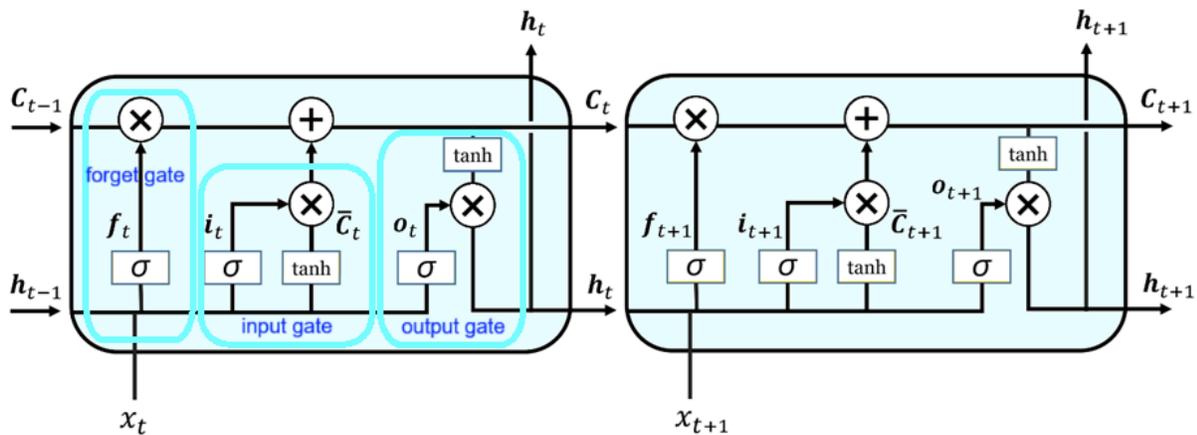

**Figure 17:** LSTM block diagram. (Adapted from [23])

The LSTM memory block architecture, depicted in Figure 17, contains three gates: the forget gate, the input gate, and the output gate. The forget gate determines what information from the previous cell state (referred to as $C_{t-1}$) should be discarded or preserved. In this gate, both the current input $x_t$ and previous hidden state $h_{t-1}$ are processed through a sigmoid function, $\sigma(x) = 1/(1 + e^{-x})$, resulting in the output $f_t$ denoted as:

$$f_t = \sigma\big(W_f[x_t, h_{t-1}] + b_f\big)$$

where $[x_t, h_{t-1}]$ is the concatenation of the current input and the previous hidden state, and $W_f$ and $b_f$ are the weight and bias parameters, respectively. The output value ranges from 0 to 1, with 0 representing complete forgetting and 1 representing complete retention of information. This output vector is then element-wise multiplied with the previous cell state $C_{t-1}$ to selectively erase or retain information.

The input gate controls the flow of new information into the cell state. It involves three parts: the hyperbolic tangent function $\tanh(x) = (e^x - e^{-x})/(e^x + e^{-x})$, which generates a candidate cell state vector $\bar{C}_t$ with values ranging from -1 to 1; the sigmoid function $\sigma$, which outputs $i_t$ with values between 0 and 1, deciding how much of the new candidate cell state $\bar{C}_t$ should be added to the cell state; and finally multiplying the outputs of these two functions ($i_t * \bar{C}_t$) to update the cell state $C_t$. The computations to obtain $\bar{C}_t$, $i_t$, and $C_t$ are given as follows:

$$\bar{C}_t = \tanh(W_c[x_t, h_{t-1}] + b_c)$$

$$i_t = \sigma(W_i[x_t, h_{t-1}] + b_i)$$

$$C_t = f_t * C_{t-1} + i_t * \bar{C}_t$$

Here, $(W_c, b_c)$ and $(W_i, b_i)$ refer to the weights and biases associated with the new candidate cell state and the input gate, respectively. The $*$ symbol denotes element-wise multiplication. In essence, the input gate takes both the current input $x_t$ and previous hidden state $h_{t-1}$ and determines which information from $x_t$ and $h_{t-1}$ should be integrated into the cell state $C_t$.

The third gate, called the "output gate," decides which information should be passed as the output. It takes into account the present input $x_t$, the previous hidden state $h_{t-1}$, and the recently updated cell state $C_t$. The output gate generates a vector, $o_t$, which acts as a filter on the cell state to produce the current hidden state $h_t$, which can be obtained using the following:

$$o_t = \sigma(W_o[x_t, h_{t-1}] + b_o)$$

$$h_t = o_t * \tanh(C_t)$$

where $o_t$ is the output of the output gate at the current time step, $\sigma$ represents the sigmoid activation function, and $W_o$ and $b_o$ are the weights and biases associated with the output gate, respectively. By using $o_t$ to weight the component of $\tanh(C_t)$, only relevant information is included in the hidden state.

These gates, through their activation functions and learned parameters, allow LSTMs to selectively retain, update, and output information, enabling LSTMs to address the challenge of long-term dependencies in sequential data.

### 3.5.2 Bidirectional LSTMs

Bidirectional LSTMs, proposed by [36], form an expansion of the traditional LSTM model. Two concurrent LSTM layers are integrated to process the input sequence in both forward and backward directions, as illustrated in Figure 18. In this configuration, the network can consider both past and future contexts for each time step. This is especially beneficial for tasks where understanding the context of a point in a given sequence requires information from both before and after it.

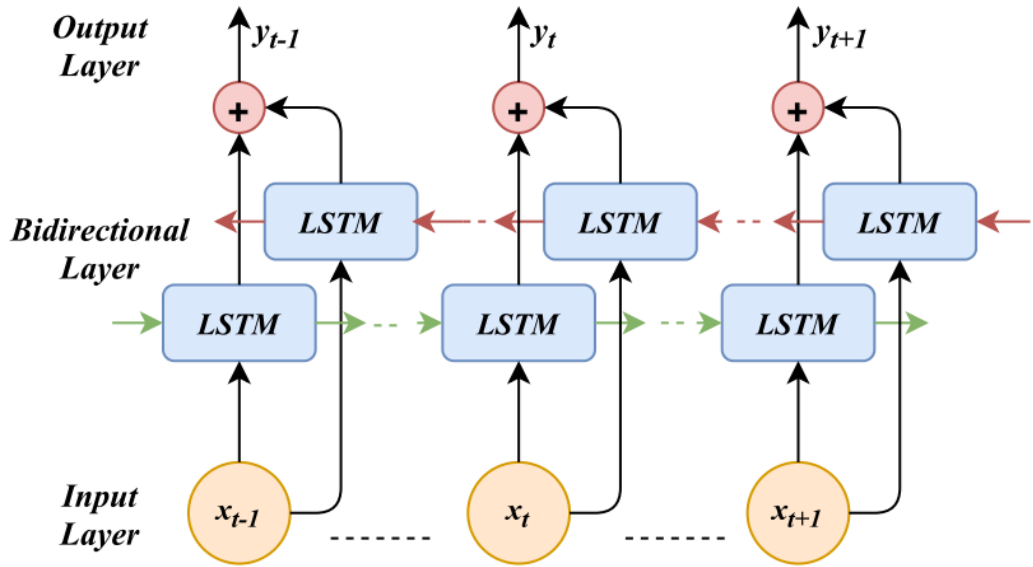

**Figure 18:** Bidirectional LSTM model architecture. (Source: [37])

### 3.5.3    Hyperparameter optimization

Hyperparameters are adjustable parameters in machine learning algorithms that control the model architecture and performance. They are not learned from the data but are set before the training process begins so that the model can be trained in the desired way. For example, the number of hidden layers in a neural network is a hyperparameter. If the number of hidden layers is too low, the model may not be able to learn the complex patterns in the data. On the other hand, if the number of hidden layers is too high, the model may overfit the training data and perform poorly on unseen data. Hyperparameter selection can have a major impact on model performance [38]. Tuning hyperparameters helps optimize the model for improved accuracy and generalization.

There are various ways to tune the parameters of machine learning models to increase their performance. While we can manually adjust hyperparameter values until we are satisfied with the results, this can be tedious and time-consuming, particularly in cases with many hyperparameters or a large number of values to consider. To address this challenge, more systematic approaches, such as grid search and random search, have been used. These methods are more efficient and effective than manual adjustments, especially for large hyperparameter search spaces that cannot be reduced using domain knowledge.

Grid search is a simple and popular technique for hyperparameter optimization. In grid search, a grid of hyperparameter values is defined and the model is then trained and evaluated on each combination of hyperparameter values in the grid. For example, if there are two hyperparameters, A and B, with three and four possible values, respectively, grid search would evaluate 3 x 4 = 12 different combinations. The combination that gives the best performance is then used in the final model. This guarantees that every possible combination of hyperparameter values will be tried, but it can be computationally expensive, especially for models with many hyperparameters, as it must train and evaluate the model on every possible combination of hyperparameter values, even if some combinations are less likely to be optimal.

Random search randomly selects combinations of hyperparameters from a predefined search space. This is in contrast to grid search, which systematically explores all possible combinations of hyperparameters. Random search can find good hyperparameter configurations without exhaustively exploring all possibilities, which saves time and resources and makes it a favorable option when the hyperparameter search space is large or when computational resources are limited. However, random search can bog down on complex models as they can take a long time to train and evaluate.

One common drawback of grid search and random search is that they evaluate each hyperparameter combination independently of the previous evaluations. This can result in inefficient allocation of time and resources as it may lead to the examination of less promising areas within the search space [39].

A more efficient approach to hyperparameter tuning is Bayesian optimization [40]. It builds a probabilistic model, which serves as a proxy for the actual objective function, based on Bayesian statistics, and uses the model to determine which set of hyperparameters to evaluate next, based on the results of previous evaluations. This helps avoid unnecessarily exploring the entire search space by focusing on the most promising regions. As a result, Bayesian optimization finds the best hyperparameters faster than grid search or random search. In this study, we use the Python library Hyperopt [41] to implement Bayesian optimization and automate the process of hyperparameter optimization.

HyperOpt offers a systematic and efficient way to search for optimal hyperparameters for our LSTM models. To use HyperOpt, we first specified the hyperparameters to be optimized and their respective search ranges. Those hyperparameters include values such as learning rates, optimizer choices, dropout rates, activation functions, bidirectional (a Boolean indicating whether to use bidirectional layers), and window sizes. These parameters control model architecture, training settings, and data processing. Table 6 shows the summary of the hyperparameter search space explored, which is detailed next.

**Table 6:** Hyperparameter search space.

| Hyperparameter | Value |
|---|---|
| learning_rate | $10^{-6} - 10^{-2}$ |
| optimizer | [ Adam, Adadelta, RMSprop ] |
| dropout_rate | $0 - 1$ |
| recurrent_dropout_rate | $0 - 1$ |
| lstm_activation | [ 'linear', 'sigmoid', 'tanh', 'relu' ] |
| output_activation | [ 'linear', 'sigmoid', 'tanh', 'relu' ] |
| bidirectional | [ True, False ] |
| window_size | [ 1, 2, …, 10 ] |
| window_size_funding | [ 1, 2, …, 10 ] |

Hyperparameter *learning_rate*: The learning rate controls how quickly a neural network updates its parameters to learn from the training data. A higher learning rate makes the network learn faster but less stably, while a lower learning rate makes the network learn stably but slower. We evaluated a range of learning rates from $10^{-6}$ to $10^{-2}$.

Hyperparameter *optimizer*: The optimizer is the algorithm used to update the weights of the model during training to minimize the loss function, which is a measure of how well the model is performing on the training data. We considered three optimizers in this study: Adam, Adadelta, and RMSprop.

Hyperparameter *dropout_rate*: The dropout rate is a regularization technique to prevent overfitting. It works by randomly dropping out the connections between the LSTM units in the same time step during training. We explored dropout rates of 0 to 1, meaning dropping out 0% to 100% of the connections.

Hyperparameter *recurrent_dropout_rate*: This parameter specifies the recurrent dropout rate, which controls the amount of dropout applied to the recurrent connections (i.e., connections between time steps) in the LSTM units. We tested dropout rates from 0 to 1.

Hyperparameter *lstm_activation*: This specifies the activation function used for the output gate of the LSTM layer. The default value is tanh, but we also experimented with other activation functions, including linear, relu, and sigmoid.

Hyperparameter *output_activation*: The output activation function transforms the network's outputs to fit the task. For regression tasks, like the one explored in our study, it is common to use a linear activation function, but we also explored relu, sigmoid, and tanh.

Hyperparameter *bidirectional*: We added this Boolean hyperparameter to the search space to represent the choice between using standard or bidirectional LSTMs so that we could explore which LSTM architecture works best for our task. We tested its values of "True" indicating bidirectional LSTMs and "False" for standard LSTMs.

Hyperparameter *window_size*: This input sequence length parameter defines the length or size of the sliding window that the model considers when making predictions. It governs the number of time steps in launch dates, failure dates, and lifetimes that the model takes into account. We tested window sizes ranging from 1 to 10.

Hyperparameter *window_size_funding*: This defines the sequence length for the funding-related features, such as the total R&D expenses, national defense R&D expenses, space R&D expenses, and NASA's budget. We also investigated from 1 to 10 time steps.

After defining the hyperparameters to be optimized, an objective function is created to guide the iterative search for the best hyperparameters. The function starts by sampling a set of hyperparameters from the defined search space and then processes the data and configures the model based on the chosen hyperparameters, making both ready for training. Upon training completion, the function calculates the loss value on the test data, which needs to be minimized, to evaluate the model's performance with those hyperparameters. This process repeats for another set of hyperparameter values until reaching a specified maximum number of evaluations or trials, iteratively updating the best hyperparameters. The results of each trial are recorded to keep track of the hyperparameter combinations and their corresponding loss values, ultimately finding the best set of hyperparameters that leads to the minimum loss value.

### 3.5.4    Time-based data splitting

The performance of a machine learning model can vary significantly depending on how the data are split [42]. To avoid overfitting and achieve good model generalization performance, it is crucial to have a balanced training and test set. We used two data split ratios, 75%-25% and 85%-15%, to split the data based on the temporal order of the data. Also, to validate the model, the training set was split into a training and validation set using the same ratio. Therefore, in the case of a 75-25 data split, the final datasets were divided into 56% for training data (75% of 75% equals 56%), 19% for validation data (75% of 25% equals 19%), and the remaining 25% for testing. With an 85-15 split, the datasets were allocated at 72% for training data (85% of 85% equals 72%), 13% for validation data (85% of 15% equals 13%), and 15% for testing. During the training process, the model was exposed to both the training and validation sets, which helped it learn the patterns in the data. Following training, its performance was assessed on a separate test set to evaluate the model's generalization ability. This test set comprised unseen data, ensuring the model was not simply memorizing the training data. This evaluation helped assess how well the model would generalize to new data. We evaluated the model's performance on each split ratio and adopted the one that resulted in better model performance.

### 3.5.5 Performance metrics

In the training phase, the Mean Square Error (MSE) on the cross validation set was used to evaluate the performance of the trained model. MSE calculates the average squared difference between the values predicted by a model and the actual observed values (Eq. 9). It is a common choice for regression problems during the model development and hyperparameter tuning phases. To ensure that the models generalize well to new data, we then tested the model on a held-out test set using the Root Mean Square Error (RMSE), Eq. 10. This metric is measured in the same units as the target variable, which makes it easy to interpret and compare prediction errors on the original data scale.

$$MSE = \frac{1}{n} \sum_{i=1}^{n} (y_i - \hat{y}_i)^2 \qquad (9)$$

$$RMSE = \sqrt{MSE} \qquad (10)$$

Here, $n$ is the number of samples, $y_i$ is the actual value, and $\hat{y}_i$ is the predicted value for the $i$-th sample.

### 3.6 Benchmark model

Establishing a minimum prediction performance level helps assess whether more complex models offer a significant advantage in terms of prediction accuracy. To achieve this, we used a regression-based model, based on our previous work [9], as a benchmark to compare the performance of our LSTM models. This benchmark model predicts log-lifetime of spacecraft as a function of time and additional predictors, combining both log-transformed and linear input features. Specifically, the model employs a partially log-linear structure, expressed as follows:

$$\log_2(L_t) = \alpha + \beta_0 t + \beta_1 \log_2(X_{1t}) + \sum_{i=2}^{k} \beta_i \log_2(X_{it}) + \sum_{j=k+1}^{n} \beta_j X_{jt} + \epsilon$$

This is achieved by applying the logarithm to the general form of Moore's law and incorporating additional relevant features into the resulting log-transformed equation. Equivalently, we regressed the log-transformed spacecraft lifetimes $L_t$ on the set of input features. In this context, $t$ corresponds to the "Launch date," covering the time span from 1959 to 1999 during which all spacecraft launched were no longer active (except Voyager 1 and 2 launched in 1977, whose projected lifetime figures stem from publicly available estimates). $X_{1t}$ represents the "Launch mass," while $X_{it}$, for $1 < i < k+1$, denotes a moving average over a specified window size in years computed on the variables that represent yearly R&D investments and NASA's annual budget, such as "Total R&D," "National Defense R&D," "Space R&D," and "NASA Budget," for the year of the launch date ($t$). $X_{jt}$, for $j = k+1, \dots, n$, represents dummy variables from one-hot encoded categorical variables, encompassing "Destination," "Type of contact," and "Country of MFR." Parameter fitting for $\alpha$, $\beta_0$, $\beta_1$, $\beta_i$, and $\beta_j$ was conducted using the ordinary least squares method.

To achieve the most reliable and accurate predictions possible, a two-step approach was employed to address the issue of potential collinearity among the independent variables. First, a Pearson correlation coefficient heatmap was generated (Figure 19) to visually identify any highly correlated variables. The Pearson correlation coefficient measures the linear correlation between variables, with values ranging from -1 to 1. The highest correlation was found between "log_NationalDefenseR&D" and "log_TotalR&D" with a value of 0.9. The second-highest correlation was observed between "log_NationalDefenseR&D" and "Launch date," as well as between "log_TotalR&D" and "Launch date," both yielding a correlation coefficient of 0.8. Given that "Launch date" exhibited a more pronounced correlation with the dependent variable "log2(Lifetime) " than both "log_TotalR&D" and "log_NationalDefenseR&D," we chose to retain "Launch date" and excluded it from the second step. Another notable correlation was between "log_NASABudget" and "log_SpaceR&D" with a value of 0.77.

Secondly, a feature selection process was implemented to exclude variables with high collinearity so as to identify the optimal combination of variables for the model. This process involved fitting the model with various combinations of the highly correlated variables for each window size from 1 to 20. To ensure robust evaluation and reduce overfitting, a 5-fold cross-validation approach was employed during the model fitting process. This technique involved splitting the data into five folds, training the model on four folds and evaluating its performance on the remaining fold. This process was repeated five times, ensuring each data point was used for both training and validation. The models were then evaluated based on prediction error (measured by the mean RMSE of the cross-validation) and multicollinearity (measured by variance inflation factor, VIF). There is no single VIF threshold that definitively indicates multicollinearity and its negative impact on prediction models. For our analysis, we chose a VIF cutoff of 5, following a commonly used guideline in many research fields. The model selection prioritized a configuration that achieved the lowest RMSE while maintaining VIFs below the threshold of 5, aiming to minimize prediction error while mitigating the effects of potential collinearity among the variables. This ensures that the model coefficients remain reliable, leading to more accurate and generalizable predictions for unseen data.

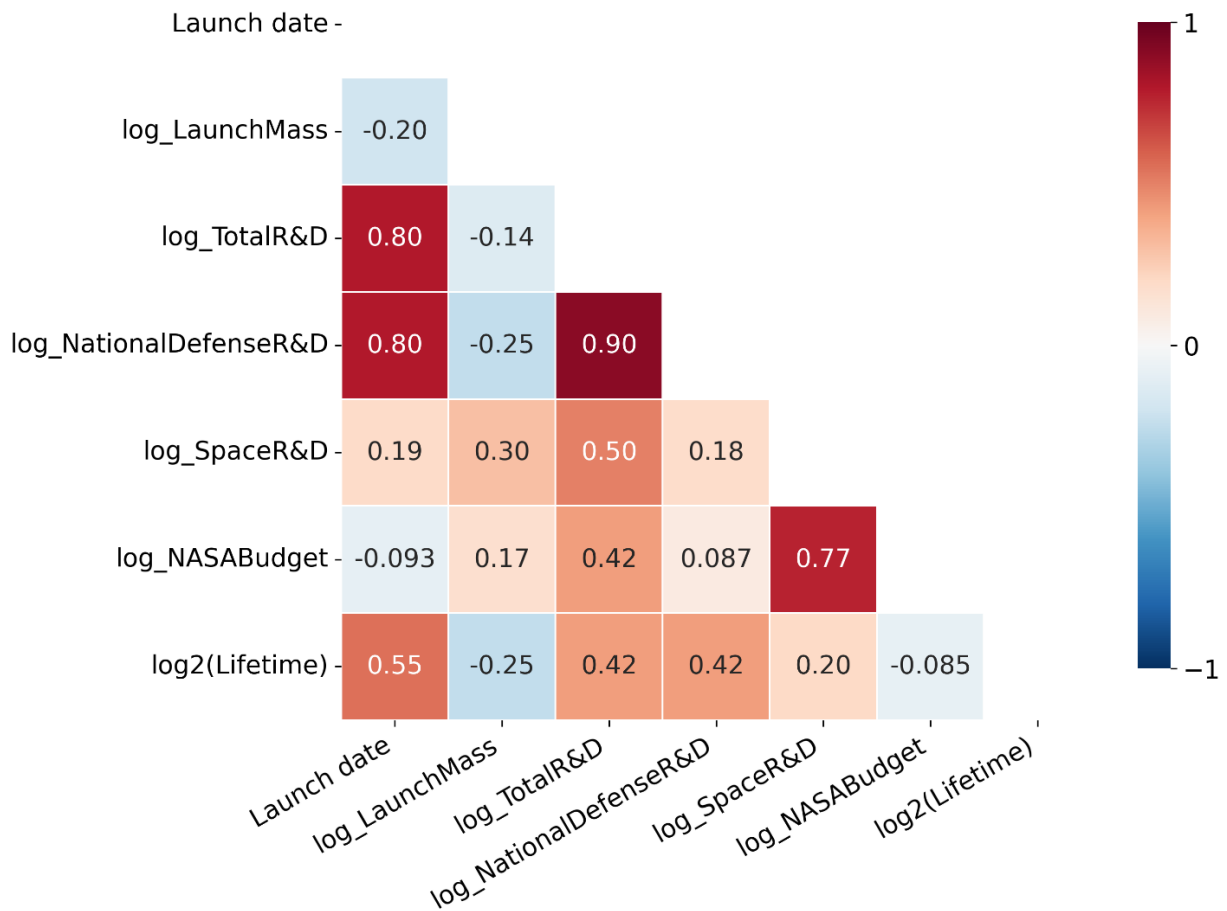

**Figure 19:** Correlation matrix heatmap. Darker color indicates stronger correlation.

The results, including the specific variable combinations, corresponding lowest mean RMSE, and associated window size, are presented in Table 7. Based on the analysis of the mean RMSE of cross-validation, model 13 with a window size of 10 was chosen for the analysis. This model achieved a mean RMSE of 1.7594 and maintained VIF values for all variables below the threshold of 5. Table 8 displays the VIF values associated with the model's variables.

**Table 7:** Fitting results.

| Model No. | $X_{it}$ Variables | Lowest Mean RMSE | Best Window Size |
|-----------|--------------------|------------------|------------------|
| 1 | "NASA Budget," "National Defense R&D," "Space R&D," "Total R&D" | 2.3161 | 5 |
| 2 | "NASA Budget," "National Defense R&D," "Space R&D" | 2.2725 | 10 |
| 3 | "NASA Budget," "National Defense R&D," "Total R&D" | 1.7942 | 9 |
| 4 | "NASA Budget," "Space R&D," "Total R&D" | 2.3666 | 6 |
| 5 | "National Defense R&D," "Space R&D," "Total R&D" | 1.8474 | 8 |
| 6 | "NASA Budget," "National Defense R&D" | 1.8369 | 10 |
| 7 | "NASA Budget," "Space R&D" | 2.1723 | 10 |
| 8 | "NASA Budget," "Total R&D" | 2.1792 | 10 |
| 9 | "National Defense R&D," "Space R&D" | 1.8790 | 10 |
| 10 | "National Defense R&D," "Total R&D" | 1.8692 | 10 |
| 11 | "Space R&D," "Total R&D" | 2.0933 | 10 |
| 12 | "NASA Budget" | 2.3418 | 10 |
| **13** | **"National Defense R&D"** | **1.7594** | **10** |
| 14 | "Space R&D" | 2.3385 | 10 |
| 15 | "Total R&D" | 2.2657 | 10 |

**Table 8:** VIF values of the best model.

| Variables | VIF |
|-----------|-----|
| Constant | 0.000 |
| "Launch date" | 4.838 |
| log2("Launch mass") | 2.544 |

| log2("National Defense R&D") | 4.152 |
|---|---|

## 3.7 Model specification and implementation

The model was built using Python's Keras functional API, which runs on top of TensorFlow. The API allowed us to build an LSTM model with multiple inputs. Figure 20 shows a high-level diagram of the model architecture. The model begins with various input layers that take different types of data as inputs. These include input layers for numerical features and input layers for categorical features. Next, the model applies two LSTM layers to process the sequential numerical features that have been formatted to align with the selected *window_size* and *window_size_funding*. Depending on the value of the *bidirectional* hyperparameter, which determines whether bidirectional LSTMs are used, these layers can be either standard LSTMs or bidirectional LSTMs. The number of neurons in these layers is determined by the shape of the input data. The dropout rate, controlled by the *dropout_rate* hyperparameter, is employed to regularize the input connections within the LSTM layers to prevent overfitting. Additionally, the *lstm_activation* hyperparameter specifies the activation function to be used in the LSTM layers, while *recurrent_dropout_rate* dictates the dropout rate applied to the recurrent connections within these layers. Following the LSTM layers, the model applies dense layers to transform the LSTM outputs. Each LSTM output is connected to a dense layer and is reshaped to match the desired output format. For each categorical feature, such as "Destination," "Type of contact," and "Country of MFR," an embedding layer was added after the input layers to transform the feature into a unique vector that can be learned by the model. Considering computational efficiency, we set the length of the embedding vectors to approximately half the number of unique categories. Subsequently, the model uses a concatenation layer to concatenate the embedding vectors from different categorical features together with the non-sequential numerical inputs ("Launch mass") and the outputs from the LSTM and dense layers to produce a composite feature vector, which is then passed through a batch normalization layer to stabilize and accelerate the training process. Lastly, a one-unit dense layer as an output layer is added to take the output of the batch normalization and generate the predictions. The specific activation function applied to the output layer is determined by the selected *output_activation* hyperparameter.

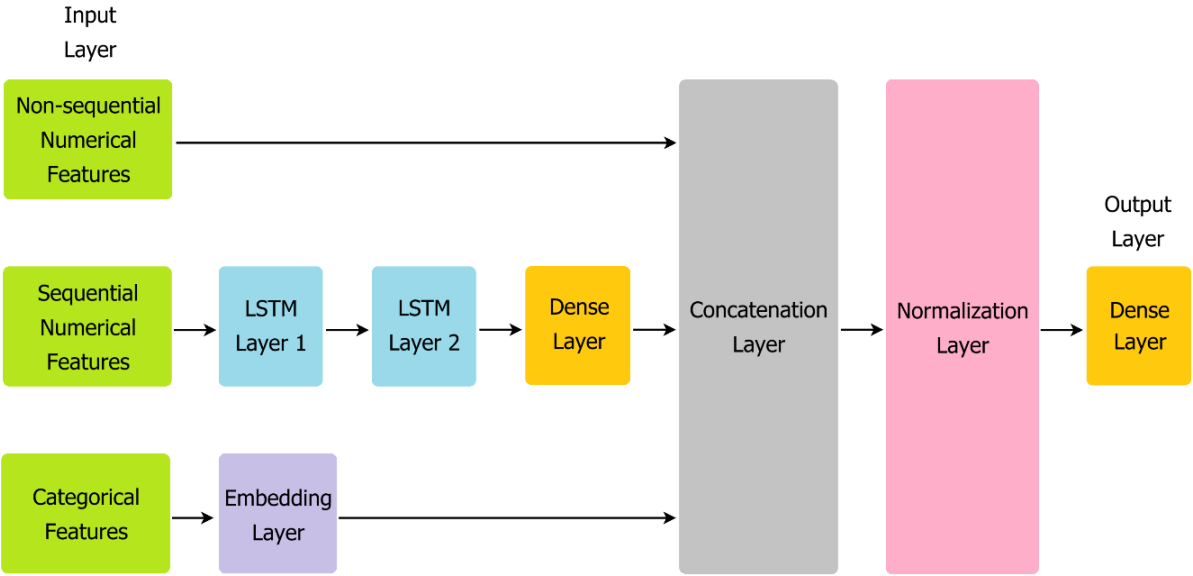

**Figure 20:** Model architecture. Input layer blocks are in green.

The model was trained using the training data, and its performance was monitored on the validation set. The training and validation data were provided to the model in batches, with each batch containing *batch_size* data points. The batch size controls how often the network's weights are updated. We assessed the impact of various batch sizes: 32, 64, 96, and the full size of the training set, on the model performance. We set the number of epochs or iterations the model will be trained on the training dataset to 5,000. In each epoch, the model updates its internal parameters to minimize the loss function, which measures the error between the model's predictions and the actual target values and was set to Mean Squared Error (MSE). The model's performance and training process were impacted by hyperparameters. Our LSTM-based model was fine-tuned through hyperparameter optimization and iterative trials. As described previously, 9 hyperparameters of the model are to be tuned while the model is being trained.

We used two callback functions, EarlyStopping and ModelCheckpoint, to prevent overfitting and monitor the model training progress. EarlyStopping monitors the validation loss and stops training if the loss does not improve for a specific number of consecutive epochs, which we set to 1,000. ModelCheckpoint records the loss and saves the model with the best validation loss. Following training, the model was evaluated on a separate test set to assess its performance on unseen data using the RMSE metric.

## 4.      Results and analysis

In this section, we evaluate the predictive performance of LSTM models trained with varying data splits and batch sizes within each modeling phase. As stated earlier, Phase 1 focused on time-only modeling, where the inputs consisted of failure times and lifetimes during the failure time-based Stage 1 and launch times during the launch time-based Stage 2. Phase 2 extended this baseline to conduct time-plus modeling by incorporating additional predictors such as NASA budgets, R&D investments, spacecraft mass, and mission characteristics. Each phase involved using the STETI two-stage modeling process, beginning with failure time-based training and transferring the predicted outputs to train the launch time-based models. We selected the best launch time-based model from Stage 2 of each modeling phase based on predictive accuracy, designating them as the best time-only model and best time-plus model, respectively. These two models were then compared to identify the overall top-performing model. Finally, we compared the performance of this top model with the benchmark regression model and analyzed its predictive behavior under various scenarios to gain insights into its adaptability.

## 4.1 Evaluation of predictive performance of LSTM models across the two phases

Table 9 presents the RMSE values for various configurations of LSTM models trained in Stage 1 of the time-only modeling phase. These models were trained using lifetimes and failure dates as input features. For comparison, results from models trained with default hyperparameters are also included. As shown in the table, models under the "Tuning" setting consistently achieved lower RMSE values across all four batch sizes and both data split ratios. Notably, the model trained with hyperparameter tuning, an 85:15 train/test split ratio, and a batch size of 32 achieved the lowest RMSE of 0.1878.

Table 10 summarizes the RMSE values for various configurations of LSTM models trained in Stage 2 of the time-only modeling phase. Among all configurations, the model trained with a 75:25 split ratio and a batch size of 96 stood out with the lowest RMSE of 1.5736 and was therefore chosen as the best time-only model. Table 11 displays the optimal hyperparameter settings for this model.

**Table 9:** Comparison of RMSE for failure time-based LSTM models in Stage 1 of the time-only modeling phase across varying hyperparameter settings, data splits, and batch sizes.

| Hyperparameter Setting | Training/Test Split Ratio | RMSE Across Different Batch Sizes | | | |
| --- | --- | --- | --- | --- | --- |
| | | 32 | 64 | 96 | Training Set Size |
| Tuned | 75:25 | 0.4067 | 0.4137 | 0.3065 | 0.46 |
| | 85:15 | **0.1878** | 0.2346 | 0.2487 | 0.2461 |
| Default | 75:25 | 0.7357 | 0.9762 | 1.7637 | 1.5774 |
| | 85:15 | 0.5681 | 0.6354 | 1.0639 | 0.8457 |

**Table 10:** Comparison of RMSE for launch time-based LSTM models in Stage 2 of the time-only modeling phase across varying hyperparameter settings, data splits, and batch sizes.

| Training/Test Split Ratio | RMSE Across Different Batch Sizes | | | |
| --- | --- | --- | --- | --- |
| | 32 | 64 | 96 | Training Set Size |
| 75:25 | 1.9176 | 1.9142 | **1.5736** | 1.6103 |
| 85:15 | 1.9073 | 1.7049 | 1.9072 | 1.7862 |

**Table 11:** Hyperparameter results for the best time-only model.

| Hyperparameter | Best Value |
| --- | --- |
| learning_rate | 0.0053 |
| optimizer | RMSprop |
| dropout_rate | 0.3717 |
| recurrent_dropout_rate | 0.7736 |
| lstm_activation | linear |
| output_activation | linear |

| bidirectional | True |
|---|---|
| window_size | 10 |

Phase 2 extended the time-only phase by incorporating additional predictors to perform time-plus modeling. Table 12 shows the RMSE values for different configurations of LSTM models trained in Stage 1 of the time-plus modeling phase. Among these configurations, the model trained with hyperparameter tuning, an 85:15 train/test split, and a batch size of 32 achieved the best performance, with an RMSE of 0.4925.

Table 13 presents the RMSE values for various configurations of LSTM models trained in Stage 2 of the time-plus modeling phase. The results show that the model trained with an 85:15 split ratio and a batch size of 96 achieved the lowest RMSE (0.9448) and was therefore selected as the best time-plus model. Table 14 lists the optimal hyperparameter settings for this model.

**Table 12:** Comparison of RMSE for failure time-based LSTM models in Stage 1 of the time-plus modeling phase across varying hyperparameter settings, data splits, and batch sizes.

| Hyperparameter Setting | Training/Test Split Ratio | RMSE Across Different Batch Sizes | | | |
|---|---|---|---|---|---|
| | | 32 | 64 | 96 | Training set size |
| Tuned | 75:25 | 1.2254 | 1.0909 | 0.9549 | 0.6124 |
| | 85:15 | **0.4925** | 0.933 | 0.8529 | 0.6345 |
| Default | 75:25 | 1.1833 | 1.2095 | 1.0225 | 0.9068 |
| | 85:15 | 0.9872 | 1.004 | 0.9785 | 1.0066 |

**Table 13:** Comparison of RMSE for launch time-based LSTM models in Stage 2 of the time-plus modeling phase across varying hyperparameter settings, data splits, and batch sizes.

| Training/Test Split Ratio | RMSE Across Different Batch Sizes | | | |
| --- | --- | --- | --- | --- |
| | 32 | 64 | 96 | Training Set Size |
| 75:25 | 1.2881 | 1.2545 | 1.2816 | 1.2238 |
| 85:15 | 0.9738 | 1.0057 | **0.9448** | 1.0266 |

**Table 14:** Hyperparameter results for the best time-plus model.

| Hyperparameter | Best Value |
| --- | --- |
| learning_rate | 0.1214 |
| optimizer | RMSprop |
| dropout_rate | 0.8518 |
| recurrent_dropout_rate | 0.8928 |
| lstm_activation | relu |
| output_activation | linear |
| bidirectional | FALSE |
| window_size | 1 |
| window_size_funding | 2 |

We also assessed the performance of the best time-only and time-plus models in predicting spacecraft lifetimes. Figure 21 provides a visual comparison of the predicted and actual lifetimes, offering insights into each model's predictive accuracy. The RMSE values shown in the legend measure the prediction errors for each model. In the figure, the sky blue line represents the actual lifetimes, displayed on a logarithmic scale for each spacecraft. The orange line shows the predictions from the best time-only model, which achieved an RMSE of 2.6152, while the blue line represents the predictions from the best time-plus model, which achieved a lower RMSE of 2.0626. Based on its superior predictive performance, the best time-plus model was designated as the overall optimal model.

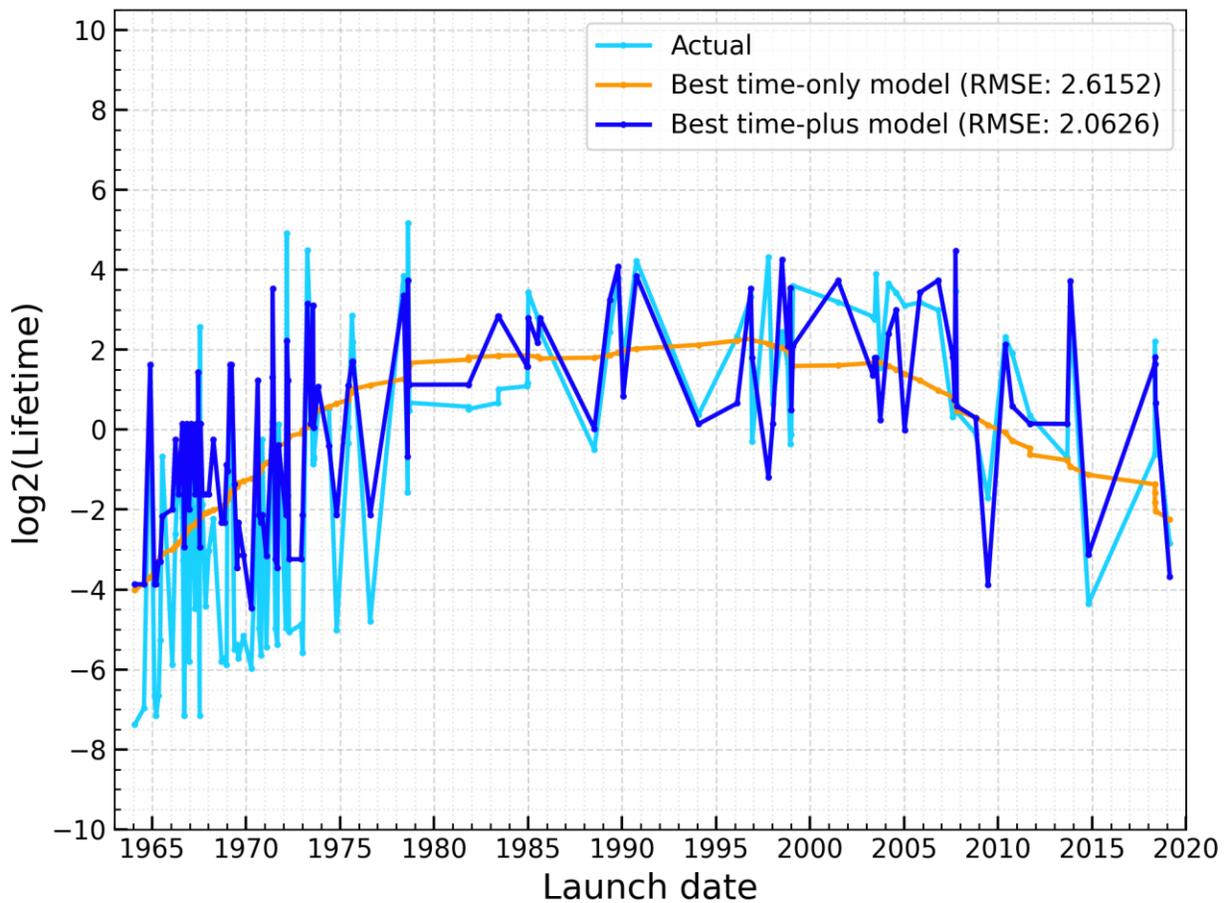

**Figure 21:** Comparison of the predictive performance of the best time-only and best time-plus models, showing that the best time-plus curve (dark blue) tracks the actual data, indicating better predictive capability.

## 4.2 Comparison of the optimal LSTM model with the benchmark model

We compared the predictive performances of the top-performing LSTM model and the benchmark multi-variate regression model described earlier on the unseen test data. Figure 22 compares the predicted and actual $\log_2$(Lifetime) values, illustrating the superior generalization ability of the LSTM model. This superiority is evident from the lower RMSE value and the visually closer alignment of the blue line (LSTM predictions) with the sky blue line (actual lifetimes), compared to the orange line (benchmark predictions). This closer alignment suggests that the optimal model more effectively captured the underlying trends in the test data, leading to more accurate predictions for the held-out spacecraft lifetimes.

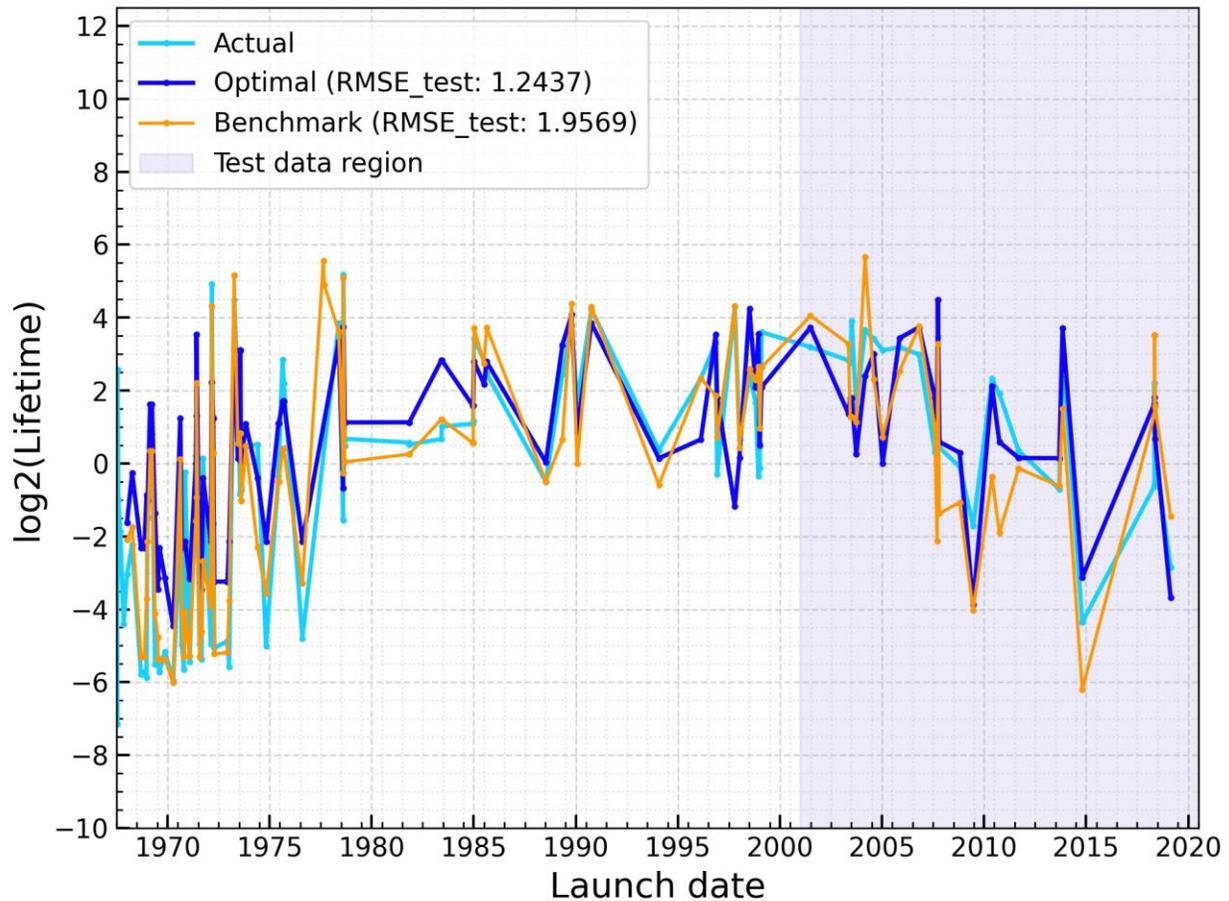

**Figure 22:** Comparison of benchmark and optimal model predictions against actual spacecraft lifetimes. The "Optimal" curve is better than the benchmark curve, as quantified by a smaller RMSE value.

### 4.3    Analysis of the optimal model's predictions in different scenarios

We delved deeper into the optimal model's predictions in various scenarios to better understand its capabilities. As the actual lifetime data for future launches is, naturally, unavailable, we investigated the model's behavior when presented with hypothetical launch scenarios. To create these scenarios, we utilized the final data point in the dataset, representing HAKUTO-R M1, a 1,000 kg lunar lander launched by Japan on December 11, 2022. Based on this reference point, we generated predictions for our analysis by varying specific features of the spacecraft. Examining these predictions allowed us to assess the model's ability to extrapolate beyond the existing data and potentially uncover trends that might emerge in its projections.

### 4.3.1    Impact of launch mass on lifetime predictions

This section explores the relationship between a spacecraft's launch mass and the predicted lifetime generated by the optimal LSTM model. We analyzed the effect of launch mass by holding all other input features constant. Launch mass varies across a broad spectrum, ranging from 1 kg to 50,000 kg. To visualize the impact of this variation, we generated a series of predictions based on the final data point in our dataset. Figure 23 presents these results. Red circles represent spacecraft that are still operational, while small blue circles denote predicted lifetimes for those currently functioning. The large blue circle corresponds to the predicted lifetime of the final data point in the dataset, HAKUTO-R M1, a 1,000 kg lunar lander. The cyan vertical line illustrates the predicted lifetimes when varying only the launch mass from 1 kg to 50,000 kg, with all other features held constant. The endpoints of this line, marked by grey and orange dots, represent the predicted lifetimes at the minimum and maximum simulated masses, respectively. As shown, there is a clear inverse relationship between launch mass and the predicted value. As the launch mass increases from 1 kg to 50,000 kg, the predicted value on the y-axis exhibits a downward trend. This suggests that the model learned a pattern where larger launch masses are associated with shorter lifetimes. A plausible reason for this result is the fact that the heaviest vessels were launched relatively early in the space age when their lifetimes were shorter.

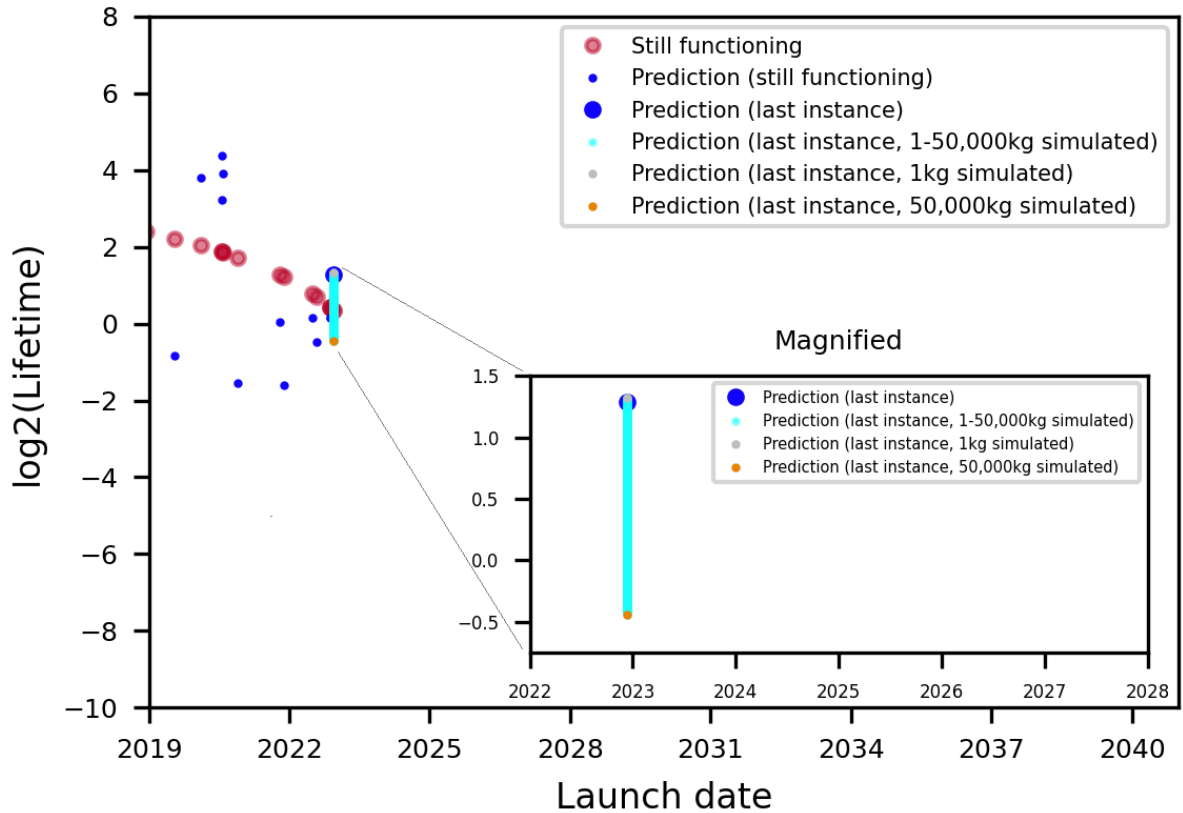

**Figure 23:** Lifetime predictions for hypothetical launch mass. This explores some hypotheticals.

### 4.3.2    Impact of country of manufacture on lifetime predictions

We examined whether the optimal model factored in country of origin, given that countries possess varying levels of expertise in specific spacecraft technologies. Figure 24 highlights how the predicted lifetime would change when assuming different countries of manufacture. Even though all other input features were held constant, the predicted lifetime went up and down depending on the hypothetical country of origin. This suggests that the model incorporates the country of manufacture as a factor influencing spacecraft lifetime.

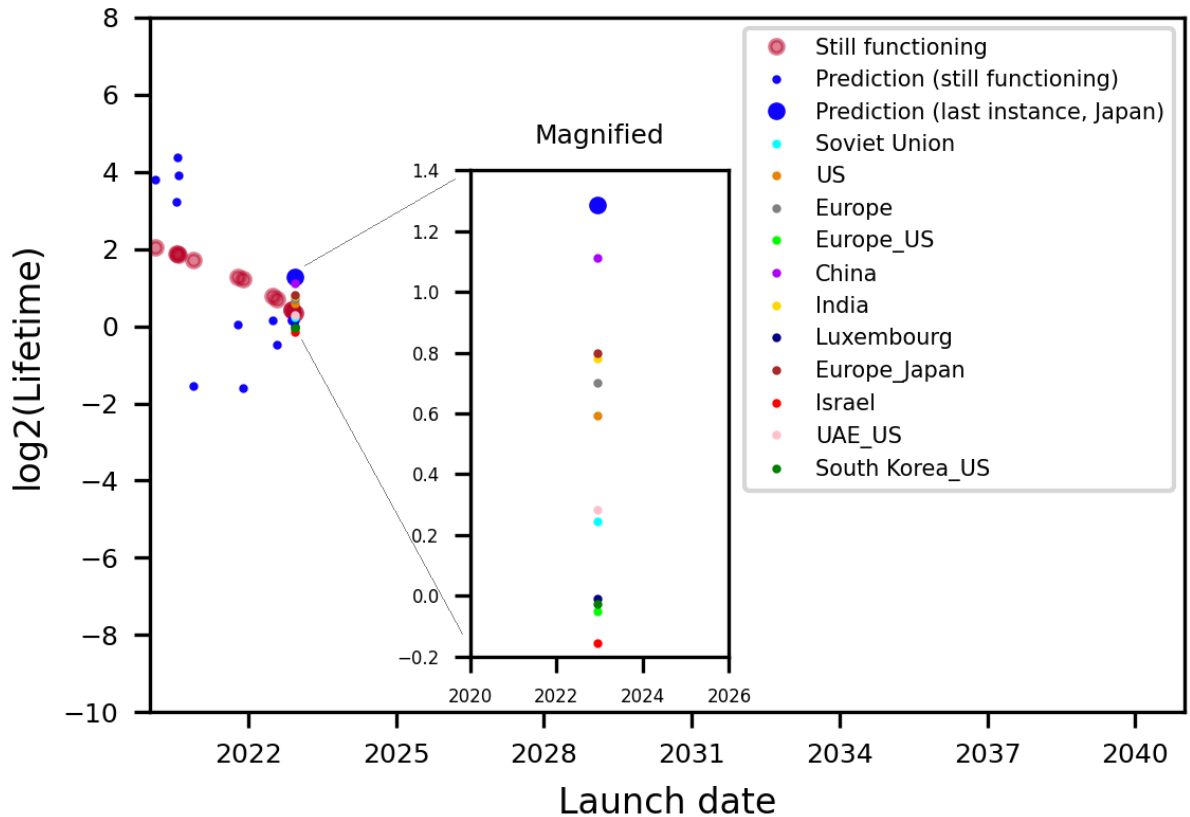

**Figure 24:** Lifetime predictions for hypothetical country of manufacture.

### 4.3.3    Impact of destination on lifetime predictions

Different destinations present unique challenges, and thus, reaching different destinations requires different functionalities. Figure 25 visualizes how changing the destination, while holding all other features constant, affects the model's predictions. As we can see from the graph, the predicted value fluctuates significantly depending on the hypothetical destination. This implies that the model considers the mission target when making predictions. A spacecraft's intended destination would likely influence its design and ultimately its performance, so this result is not surprising.

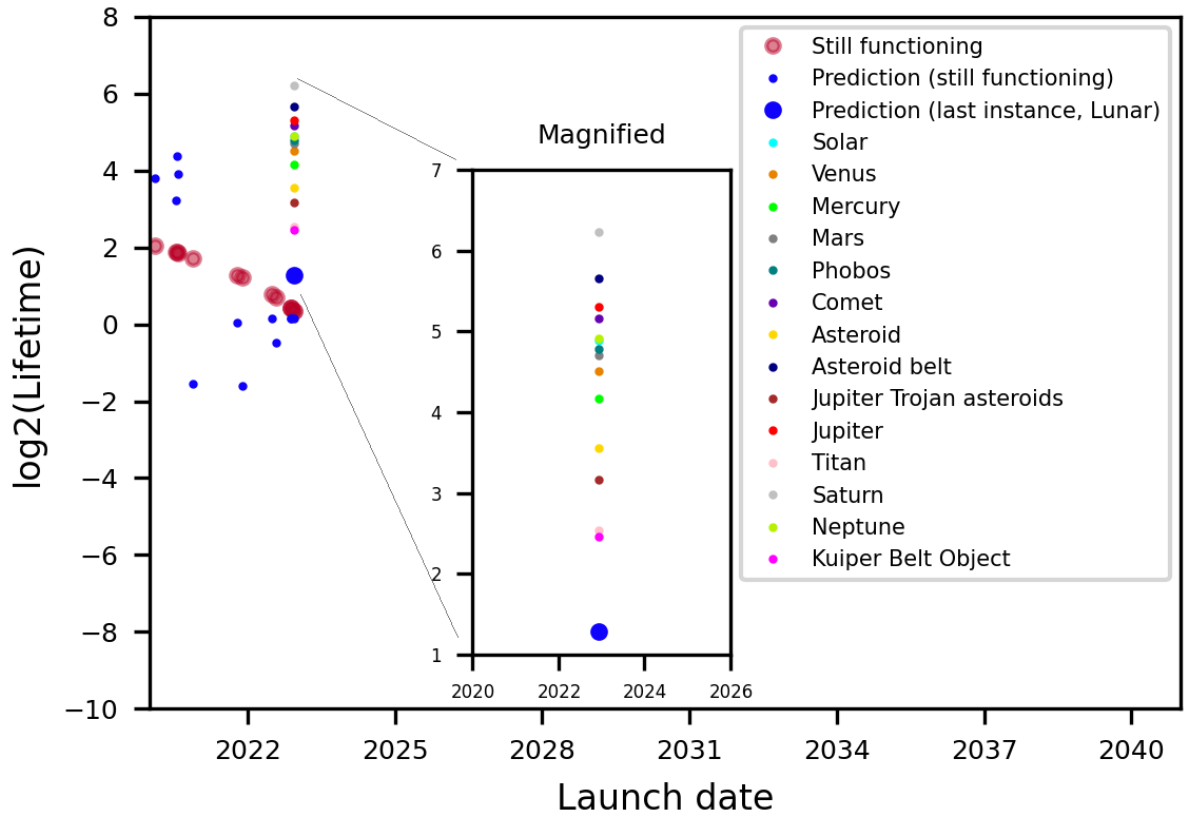

**Figure 25:** Lifetime predictions for hypothetical destinations.

### 4.3.4    Impact of type of contact on lifetime predictions

We explored how the "Type of Contact," the way a spacecraft interacts with its destination, influenced the predictions according to our model. Figure 26 illustrates how the predicted lifetime would change if this same spacecraft were tasked with alternative mission objectives on the Moon present in the data, including flyby, orbit, impact, successful landing (soft landing), unsuccessful landing, and a combination of orbit and unsuccessful landing for the spacecraft that exhibit continued operation of their orbiters despite failures of their landers during landing attempts. As the graph reveals, the predicted value fluctuates depending on the hypothetical type of contact, suggesting the model considers the mission objective beyond just the destination when making predictions.

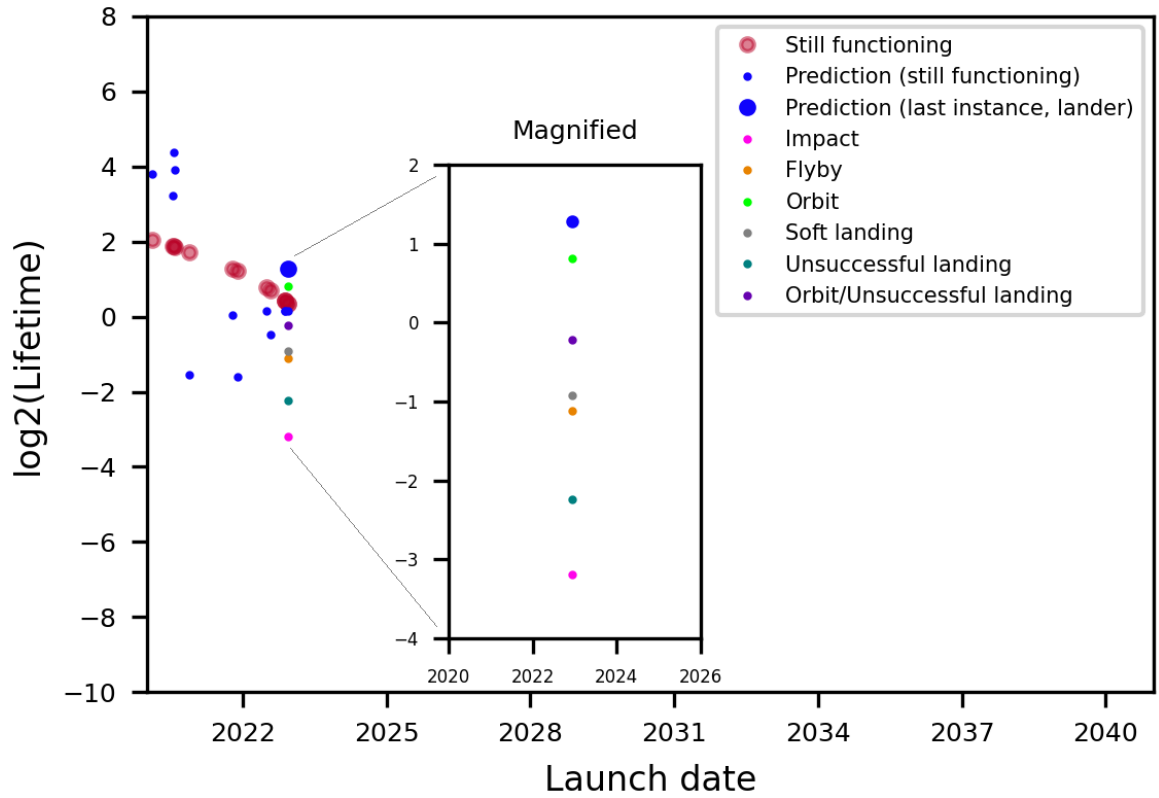

**Figure 26:** Lifetime predictions for hypothetical type of contact.

## 5. Discussion and conclusions

### 5.1 Contributions to understanding deep space exploration vessel lifetimes

This study has contributed to enhancing the effectiveness of technology forecasting in the space exploration sector.

First, a thorough and up-to-date review of existing quantitative trend extrapolation techniques applied in technology forecasting research was undertaken. This review provided a foundational understanding of how these methods have been utilized across diverse technological domains, offering valuable insights that can be leveraged to improve forecasting models for space exploration technologies.

Second, an approach was presented to transform the standard start-time based exponential function into an end-time based function. This approach allows for the effective utilization of failure data from relatively recent launch years, which is valuable for forecasting future spacecraft lifespans but present a challenge to use due to the issue of recent-year data bias.

Third, building upon this foundation, a novel predictive model of expected spacecraft lifetime as a function of start time and other spacecraft properties was developed. This model efficiently uses available lifetime data to forecast the expected lifespan of future spacecraft, leading to more accurate forecasting capabilities.

Fourth, a dataset designed for technology forecasting specifically for spacecraft used for exploration of objects in deep space was developed. While spacecraft lifetimes might serve as a proxy for technological advancement in spacecraft in general, it focuses directly on a key reason for building spacecraft physical travel to and exploration of extraterrestrial worlds. This is surely a major part of the natural appeal of spacecraft in human imagination. This dataset involved the identification of potentially relevant predictors that may influence spacecraft lifetimes, providing a foundation for future forecasting efforts.

Fifth, this research has demonstrably improved the effectiveness of technology forecasting in space exploration by building and applying a finely tuned model that integrates machine learning algorithms, specifically LSTMs, with an augmented version of Moore's law that incorporates additional influencing factors, enabling it to capture complex, long-term trends and reflect growth patterns beyond simple time-based extrapolation. The resulting model minimizes the discrepancy between predicted and actual spacecraft lifetimes through out-of-sample evaluation, resulting in more reliable forecasts.

## 5.2 Conclusions from model development

Ultimately, this study is expected to contribute to the development of decision-making tools. Policymakers, researchers, and other stakeholders within the space sector need such tools to plan and implement space missions and programs. This will support improved resource allocation and mission design.

Our analysis demonstrates that an optimized LSTM model outperforms a regression benchmark model. This superiority seems to arise from the LSTM's ability to express prediction trends that are more complex than parameterized regression functions, like exponential trend functions known as Moore's laws. This enables more accurate predictions for unseen data, enabling more reliable predictions of spacecraft lifetimes.

Furthermore, integrating relevant features associated with a spacecraft's lifespan with the exponential growth curve (generalized Moore's law) significantly improves prediction accuracy. This highlights the importance of considering factors such as launch weight, target destination, and country of origin. By incorporating these features, we create a more robust and reliable model for predicting spacecraft lifetimes.

This study also reveals that specific features can have a substantial impact on predictions. These include the spacecraft's launch weight, its target destination, and its country of origin. Understanding the influence of these features should allow building even more dependable and informative models in the future.

Finally, the research underscores the importance of hyperparameter tuning in achieving optimal model performance. While tuning the hyperparameters demonstrably improves model accuracy through a lower RMSE, it also increases training time. This highlights the importance of finding a balance between achieving the best possible performance and the computational resources required to reach it.

In conclusion, through leveraging the strengths of LSTMs to mine the effects of the passage of time as well as other relevant spacecraft attributes, and applying hyperparameter tuning, we were able to predict the lifespans of future spacecraft with greater confidence.

## 5.3     Lessons for time series forecasting and cross-domain applications

Forecasting spacecraft lifetimes presents an interesting challenge in time series forecasting. Unlike many financial or economic time series, spacecraft data has limited historical data due to the relatively small number of extraterrestrial missions launched. Additionally, for spacecraft lifetimes, complex and non-linear relationships may exist among factors such as mission type and launch weight. These factors can introduce significant variability into the data. Despite the barriers, leveraging LSTMs for this task and evaluating their performance against traditional regression models has provided useful perspectives on the field of time series forecasting.

We found that LSTMs are adept at tackling complex time series data and modeling non-linear patterns, making them a robust tool for time series forecasting. Spacecraft failures are often complex events with multiple contributing factors, and LSTMs' ability to learn and retain information from temporal dependencies proved crucial in this case. However, this study also highlights some limitations of LSTMs in time series forecasting. Training them can be time-consuming and challenging as it requires careful hyperparameter tuning and optimization of model architecture to achieve optimal performance, and can be computationally demanding, especially with extensive or complex datasets. Furthermore, understanding the internal operations of LSTMs still presents difficulties. The lack of interpretability makes it difficult to identify the specific factors influencing the model's predictions. Lastly, overfitting can occur with LSTMs, especially when data is limited and model design and training are not carefully considered.

The techniques developed in this analysis for forecasting spacecraft lifetimes could be adapted to other domains where data exhibits similar characteristics, particularly in situations where the outcome of interest is the time until the occurrence of a specific event. Applicable tasks, for instance, could involve predicting equipment failures in factory operations, evaluating effects of treatment protocols on disease progression and survival in healthcare, and modeling credit risk of future default in the financial sector. Moreover, the challenge of forecasting spacecraft lifetimes highlights the importance of selecting appropriate models. For example, analyzing how LSTMs compare to traditional regression models in this study highlights the need to understand the data when choosing a forecasting approach for various applications. The insights obtained from the forecasting process also can guide data collection, model development, hyperparameter tuning, and the improvement of forecasting techniques in other fields facing similar challenges.

## 5.4 Limitations and future work

Although progress has been achieved in spacecraft lifetime forecasting models, there remain limitations to address. However, these limitations also provide opportunities for future research.

One limitation of this study is the availability and scope of the data used. Factors potentially impacting spacecraft lifetime, such as research and development investments by other countries and private companies, were not considered due to constraints on obtaining often proprietary information.

An alternative modeling approach could be explored to avoid training a second LSTM on the predicted lifetimes from the first LSTM, potentially reducing the associated errors. This method would involve subtracting the predicted lifetime from the failure time to obtain the predicted launch time, resulting in pairs consisting of predicted launch time and lifetime. By comparing these predicted pairs with the actual launch time and lifetime pairs, the discrepancies could be calculated to evaluate the goodness of fit. Comparing the fit to that produced by the Stage 2 LSTMs investigated above will provide insights into the effectiveness of this approach.

This study used specific methods for model development, such as particular data split ratios and a Bayesian optimization algorithm for hyperparameter tuning. Future research could explore the impact of different data splitting methods and alternative hyperparameter optimization techniques to potentially improve model performance.

The current model uses a point forecasting approach for lifetimes. This approach provides a single value for the predicted lifetime. Future research could investigate the application of probabilistic forecasting techniques. These techniques would generate a range of possible outcomes, offering a more complete picture of the distribution of potential lifetimes, which would be beneficial for mission planning as it would facilitate a more comprehensive risk assessment associated with future spacecraft missions.

Other machine learning algorithms, such as transformer-based models, could be investigated for their applicability to spacecraft lifetime forecasting. Moreover, as new data becomes available, retraining the model is important to ensure continued prediction accuracy, a process that can be computationally expensive. Therefore, alternative model architectures with lower computational requirements or techniques for more efficient training and hyperparameter tuning could also be investigated.

Future research could incorporate eXplainable Artificial Intelligence (XAI) techniques to better understand how the model makes its predictions. Such an analysis would provide valuable insights into the relationships between the input features and the model's outputs, facilitating improved modeling efforts.

Finally, the forecasting methodologies devised hold potential for various applications in the space domain, such as predicting the health and performance of launch vehicles and forecasting trends in the participation of women in space missions [43]. The techniques could also be adapted to non-space fields such as emerging sustainable technologies, from predicting the durability of batteries in electric vehicles to refining renewable energy forecasting for solar and wind power, and smart manufacturing applications such as equipment maintenance, tool wear monitoring, and tracking the usable life of machinery and components. These advancements can foster a more efficient and sustainable future in manufacturing, energy production, and many other areas of modern society in which understanding and projecting lifetimes is an important objective.